%% file: main.tex
\documentclass[11pt]{article}

\usepackage[final]{acl}

\usepackage{times}
\usepackage{latexsym}

\usepackage[T1]{fontenc}

\usepackage[utf8]{inputenc}

\usepackage{microtype}

\usepackage{placeins}
\usepackage{makecell}
\usepackage{inconsolata}

\usepackage{graphicx}

\usepackage{algorithm}
\usepackage{algorithmic}
\usepackage{amssymb}
\usepackage{soul}
\usepackage{booktabs}

\usepackage{graphicx}
\usepackage{subcaption}

\usepackage{booktabs}
\usepackage{multirow}
\usepackage{array}

\definecolor{headerpurple}{RGB}{190,140,160}
\definecolor{rowpurple}{RGB}{155,80,115}

\usepackage[most]{tcolorbox}

\usepackage{colortbl}

\usepackage{amsmath}
\usepackage{pifont}
\usepackage{multirow}
\usepackage{array}
\usepackage{enumitem}
\usepackage{todonotes}
\usepackage{blindtext}
\usepackage{ulem}
\usepackage{xcolor}
\usepackage{wrapfig}
\usepackage{float}

\newcommand{\dataset}{\texttt{INDI-PROP{span}}}
\newcommand{\datasetPTC}{\texttt{INDI-PROP}}
\newcommand{\model}{\texttt{ReCoT}}
\newcommand{\PTC}{\texttt{PTC}}

\title{

Reasoning for Propaganda Detection via Label-Guided CoT Optimisation

}

\author{Zeba Afroz, Harsh Vardhan, Md. Shad Akhtar \\
  Indraprastha Institute of Information Technology Delhi \\
  \texttt{\{zebaa, harsh25001, shad.akhtar\}@iiitd.ac.in}}

\begin{document}
\maketitle 
\begin{abstract}

{Propaganda detection systems usually operate as black boxes, lacking explanations. To address this, we hypothesise that persuasive techniques contain definitional characteristics with respect to their social-political setting. We introduce \model{}, a reasoning-conditioned detection framework that captures definitional cues of persuasive techniques via population-level refinement and performance-guided resampling, without requiring additional explanation annotations. On \dataset{}, \model{} achieves a macro $F1_{LCS}$ of 0.364, surpassing baselines and reaching 88\% of the SFT upper bound on Avg $F1_{LCS}$. Similarly, on \PTC{},  \model{} reaches 85\% of the SFT upper bound on Avg $F1_{LCS}$, closing much of this gap while remaining interpretable. Importantly, the analysis of the reasoning pool's convergence and update pattern for \model{} shows effective refinement of reasoning CoTs. 

 }

\end{abstract}

\section{Introduction}

\begin{figure}[t]
  \includegraphics[width=\columnwidth]{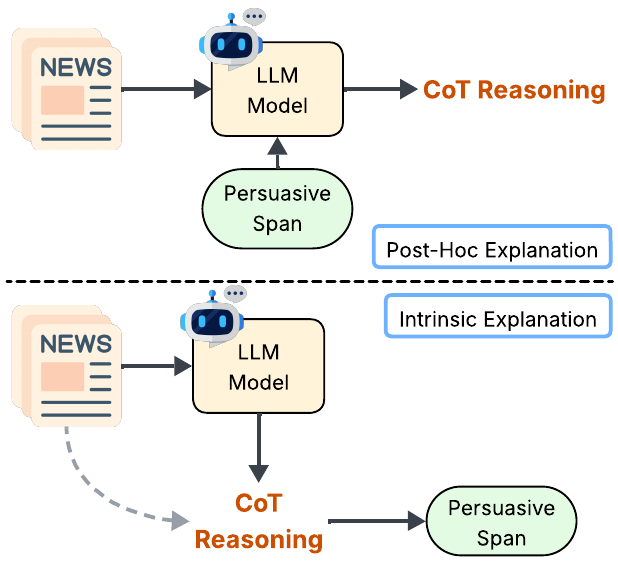}
  \caption{Conceptual comparison of Post-Hoc vs. Intrinsic Explanation frameworks. In the Post-Hoc setup (top), the Persuasive Span influences the LLM's CoT Reasoning. In the Intrinsic setup (bottom), the LLM generates CoT Reasoning directly from the input news, which then identifies the relevant Persuasive Span.}
  \label{fig: experiments}
\end{figure}

Propaganda aims to influence public opinion and create ideological biases through communicative strategies that can take textual, visual, or other forms. Unlike misinformation, propaganda often works with subtle cues to persuade. Persuasive techniques are one of the subtle strategies. Repeated exposure to these techniques can alter ideologies and normalise biased viewpoints. Hence, identifying these subtle persuasion techniques becomes crucial. Early research on propaganda detection mainly focused on basic classification tasks. The goal was to determine if a news article or paragraph contains propaganda \cite{rashkin-etal-2017-truth,barron2019proppy}. 
Building on these, later research conducted fine-grained propaganda analysis and introduced technique-wise span-level annotations \cite{da2019fine}. Further, the research advanced to multimodal and multilingual setups, as seen in SemEval workshops \cite{da-san-martino-etal-2020-semeval, dimitrov-etal-2021-semeval, dimitrov-etal-2024-semeval}. But we contend that merely identifying the specific technique is not enough, as earlier studies \cite{durmus2019exploring} have shown that readers often view information through their existing beliefs and ideological backgrounds. When propaganda aligns with a reader's viewpoint, they might see it as neutral or factual rather than manipulative. Thus, merely pointing out persuasion is not enough; an understandable explanation must accompany the detection to provide a more critical perspective for the user, enabling them to engage in critical thinking and form sound, unbiased views. 

This explainability gap is addressed by \newcite{yu-etal-2021-interpretable} by proposing a model that detects propaganda based on syntactic and semantic features. And, \newcite{zavolokina2024think} demonstrated in a user study that GPT-4-generated explanations can encourage critical thinking and increase user trust in automated systems. Further, building on these, \newcite{hasanain-etal-2025-propxplain} introduced a dataset for propaganda study in Arabic and English with explanation annotations. However, the annotations were done using post hoc reasoning, which doesn't causally guide the detection. We argue that for an explanation to be trustworthy, the reasoning must condition the decision process, ensuring an alignment between detection and its explanation. To address this, we hypothesise that persuasive techniques possess definitional cues with respect to their socio-political context. If a model can capture these cues, it can reason using them to identify persuasion, making explanation a causal part of the detection, as shown in Figure \ref{fig: experiments}. This reasoning explains the detection.

Motivated by this intuition, we propose \model{}, a reasoning-driven framework that first generates Chain-of-Thought (CoT) explanations and subsequently derives predictions conditioned on these explanations. To improve the quality of these reasoning chains (CoTs), we employ an iterative refinement strategy based on population-level cross-generalisation \cite{xu2023reprompting}, in which CoTs from multiple instances of the same persuasion technique are exchanged and reused as few-shot examples during refinement. This process encourages the CoT to capture recurring cues associated with each technique. Consequently, CoTs progressively capture all necessary cues of persuasion techniques, enabling the framework to not only identify propaganda spans but also to provide context-aware and technique-aware explanations grounded in the reasoning process itself. To ensure that captured recurring definitional cues are related to the technique and are useful, the refinement process is guided by a sampling strategy that prioritises CoTs that consistently yield measurable performance improvements across iterations. This is shown, via update-density analysis, which shows convergence of the global pool.  

Furthermore, we extend the existing \texttt{INDI-PROP} dataset \cite{afroz-etal-2026-fine} with span-level annotations for persuasion techniques. As most existing data on span-level propaganda comes from Western and Arabic contexts \newcite{}, this overlooks the socio-political contexts of countries like India. To this end, we manually annotate $3922$ spans across $20$ persuasion techniques in $1266$ news articles. 

\paragraph{Contribution:} Our contributions are as follows:
\begin{itemize}[leftmargin=*, nolistsep, noitemsep]
    \item We propose \textbf{\model{}}\footnote{Code and the subset of the dataset are available at \url{https://anonymous.4open.science/r/Prop_Span-6FF7}}, a reasoning-driven approach that employs population-based CoT sharing with iterative resampling for fine-grained propaganda detection.
    
    \item We introduce a bias-guided sampling strategy to utilize and refine reasoning over $n$ iterations. %
  
    \item We extend the existing \texttt{INDI-PROP} dataset with 3,922 span-level annotations on 20 persuasion techniques and named it \dataset{}.
\end{itemize}

\section{Related work}

Early research on propaganda detection was at the document level. \newcite{rashkin-etal-2017-truth} introduced one of the earliest datasets by compiling news articles from eight publishers and categorising them into propaganda, reliable, hoax, and satire. Building on this, \newcite{barron2019proppy} created a binary version by collapsing all non-propaganda categories into a single class. These early experiments were based on distant supervision, assuming that all articles from a given outlet share the same label. This is a highly noisy approach \cite{horne2018sampling}. Further \newcite{habernal-etal-2018-name, habernal-etal-2017-argotario, habernal-etal-2018-adapting} examined reasoning patterns and logical fallacies associated with persuasive discourse and introduced datasets targeting ad hominem and other fallacious reasoning. Although these efforts provided valuable insights into persuasive language, they did not fall within the domain of propaganda. A major advancement was made by \newcite{da2019fine} specifically for propaganda detection in news, which proposed a fine-grained annotation covering 18 persuasive techniques at the span level, reframing propaganda detection as a span-extraction task. This formulation served as the basis for the SemEval-2020 shared task \cite{da-san-martino-etal-2020-semeval}. Then, subsequent SemEval editions expanded the task to multimodal settings using memes \cite{dimitrov-etal-2021-semeval, dimitrov-etal-2024-semeval} to multilingual scenarios \cite{piskorski-etal-2023-semeval}. Despite these advances, most existing systems focus on identifying where propaganda occurs rather than explaining why a span is propagandistic. These detection systems were not interpretable. To address this gap, \newcite{yu-etal-2021-interpretable} proposed interpretable approaches that integrate syntactic and semantic features for propaganda detection. Recently, \newcite{zavolokina2024think} demonstrated in a user study that GPT-4-generated explanations can encourage critical thinking and increase user trust in automated systems. Building on these efforts, \newcite{hasanain-etal-2025-propxplain} introduced the first explanation-integrated dataset for propaganda detection in Arabic and English. Although this dataset represents an important step toward explainable propaganda detection, the explanations are generated post hoc. In contrast, our work focuses on generating intrinsic explanations alongside span-level predictions. As we argue that, prediction must be causal to reasoning. Specifically, we leverage LLMs and CoT prompting to generate reasoning without additional annotation overhead, and refine them iteratively to improve the quality and consistency of these CoTs. In addition, recent refinement methods such as Self-Refine \cite{madaan2023self}, Reflexion \cite{shinn2023reflexionlanguageagentsverbal}, and Self-Consistency \cite{wei2022chain} demonstrate the benefits of iterative refinement but rely primarily on self-feedback, memory, or ensembling. \newcite{xu2023reprompting} extends this paradigm by introducing a population-based prompt refinement strategy, and \newcite{guo2023evoprompt} is based on evolutionary algorithms. Building on this line of work, we incorporate cues of persuasion techniques to guide convergence during population-based refinement, retaining only refinements that demonstrate measurable improvements. This refinement process is implemented through resampled in-context learning. 

\section{Dataset} 

We build upon the dataset introduced by \newcite{afroz-etal-2026-fine}, which comprises 1,266 news articles covering two major socio-political events in India. We annotated each article in the \datasetPTC{} corpus with span-level labels for the existing persuasive techniques, following the definitions established by the original paper. The final corpus contains 3,922 annotated spans across 20 persuasive techniques. And we name it \textbf{\dataset{}.}

\input{tables/technique_frequency}

\subsection{Annotation}

To ensure annotation quality, we recruit annotators from diverse age ranges, ideological stances, and gender identities. In total, 9 annotators participated, aged 20-40 years, comprising 4 women and 5 men. The annotation team included research assistants, postgraduate, graduate students, and industry researchers. Additionally, two moderators were appointed to oversee and ensure the consistency of the annotations. All annotators and moderators received compensation in accordance with institutional guidelines.

\begin{figure*}[t]
  \includegraphics[width=\textwidth]{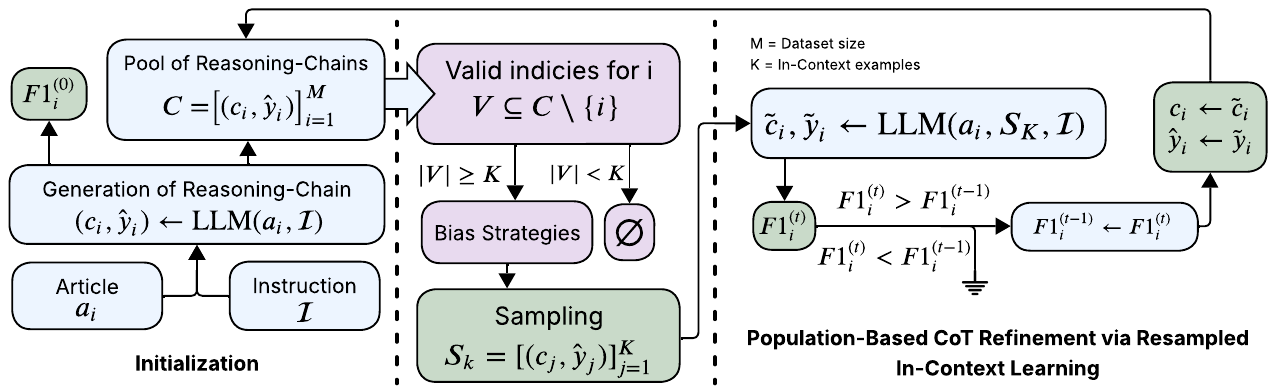}
  \caption{Overview of the proposed \model{} framework. Block 1: illustrates the initialisation stage, where initial CoT-predictions pairs are generated, Block 2: presents the sampling process \& Block 3: shows the iterative population-based refinement process via resampled in-context learning.}
  \label{fig: experiments2}  
\end{figure*}

\paragraph{Annotator Training:}The training process was organised in iterative stages to ensure conceptual alignment and consistency among annotators. 

\begin{enumerate}[leftmargin=*, noitemsep, nolistsep]
    \item \textbf{Orientation \& Pilot Annotation Phase: }The training began with an orientation session. Annotators were thoroughly familiarised with the annotation guidelines, which detailed each persuasive technique through definitions and examples. Following the orientation, annotators independently annotated an initial pilot set of articles. This pilot dataset included ambiguous cases to reveal common sources of confusion and interpretation variation at an early stage.

    \item \textbf{Annotation Refinement: }After each annotation round, meetings were conducted to compare annotations and discuss the rationale behind differing labels systematically. Inter-annotator agreement (IAA) metrics were computed after each cycle for quantitative assessment. Based on these outcomes, guidelines were refined, and discussions were iterated until satisfactory IAA scores were achieved across annotators.
\end{enumerate}

\paragraph{Final IAA Score:} 
For inter-annotator agreement (IAA), the pairwise overlap-based F1 score is used. The span overlap is calculated using the longest common substring method. The final inter-annotator agreement is calculated as the average across all annotator pairs. The IAA is 0.63, calculated for 20 articles containing 46 spans. Measuring span-level agreement is challenging as minor boundary differences reduce agreement. Moreover, the evaluation subset deliberately included ambiguous and borderline cases to stress-test annotator alignment; it represents a harder-than-average set of cases rather than a random, representative sample.

\subsection{Dataset Statistics} 

The distribution of persuasion techniques across \dataset{} shows that Smears and Assertion are the most frequent, accounting for approximately 21.7\% and 14.3\% of all persuasive spans, respectively, reflecting the prevalence of accusatory and fact-like strategies in the corpus. In contrast, Reductio ad Hitlerum is the least frequent technique, accounting for less than 1\%. Exact counts for each technique are shown in Table~\ref{tab:dataset_stats_compact}, and a detailed statistical discussion is in the Appendix \ref{Appen_Stats}.

\section{Methodology}

In contrast to the post-hoc method, where the explanation is generated as a secondary process to the predictions,  our framework directly conditions span extraction on CoT Reasoning. This allows the reasoning's performance to directly influence detection. We hypothesise that each persuasive technique has definitional cues shaped by the context in which it is implied, and to enable the CoTs to capture these cues, the CoTs were iteratively refined through population-level interactions. During refinement, only reasoning chains (i.e., CoTs) that improve span-level performance according to an acceptance criterion are retained, progressively reinforcing technique-relevant cues, as illustrated in Figure \ref{fig: experiments2}.  \model{} proceeds in two stages: (1) initialization, and (2) iterative refinement. Each article $a_i$ in the dataset $\mathcal{T} = \{a_1, a_2, \ldots, a_M\}$ is annotated with persuasive technique spans $y_i$.

\FloatBarrier

\paragraph{Initialization:} For each article $a_i$ with ground-truth labels $y_i$, we obtain an initial chain-of-thought reasoning $c_i$ and a corresponding prediction $\hat{y}_i$ using LLM in zero-shot, given instruction $\mathcal{I}$. Refer to Figure ~\ref{fig: experiments2}. The pair $(c_i, \hat{y}_i)$ forms the initial reasoning chain for article $a_i$, which serves as the starting point for subsequent refinement steps. From the initialization step, we obtain the list $C$ (Global pool), which contains all initial CoTs (Reasoning chains) for the dataset of $M$ articles.

\paragraph{Sampling (I-Bias):} \label{sec: i_bias}
\label{sec:sample_updated_bias}
We define a biased sampling strategy that favours CoTs, which have been improved in previous iterations. The primary goal is to maximise the reuse of refined reasoning. We keep $\delta$ (the bias) small and renormalise it to maintain an exploratory nature during sampling. We maintain a global probability array $\mathbf{p} \in \mathbb{R}^{M}$, initialised uniformly at the beginning for all dataset indices.  At each iteration, we aim to refine the CoT at a current index $i \in \{1, \dots, M\}$, After each refinement step, if the CoT at index $i$ yields a higher $F1_{LCS}$ score than its previous version, CoT is updated ,and the probability of index $i$ is incremented by a fixed bias $\delta = 0.02$, subject to a maximum cap of $p_{\max} = 0.1$, illustrated as \begin{math} \hat{p}_i^{(t)} = \min\left(p_i^{(t-1)} + \delta,\ p_{\max}\right) \end{math}. This will increase the likelihood of sampling the improved CoTs. For sampling CoTs, we maintain a subset $V \subseteq C \setminus \{i\}$, which denotes the set of valid indices corresponding to non-empty CoTs. $i$, denoting the current index, is excluded from the candidate pool of valid indices to avoid self-conditioning.

\begin{equation} \small
f(i, K) =
\begin{cases}
\text{Sample}\left(V,\ K,\ \mathbf{p}_V^{(t)}\right), & \text{if } |V| \geq K \\
\emptyset, & \text{if } |V| < K
\end{cases}
\label{eq:updated_bias_sampling}
\end{equation}

\noindent After increasing the bias, the global probability array is then renormalised to maintain a valid distribution. The sampling function $f(i, K)$ then draws $K$ support indices (K in-context examples) from $V$ as shown in equation \eqref{eq:updated_bias_sampling}, where $\mathbf{p}_V^{(t)}$ is the renormalized distribution restricted to $V$. If instances in the valid set are less than the value of K, which is the required number of in-context examples, then in this scenario, a null set is passed. This bias strategy ensures that CoTs that have shown consistent improvement are more frequently sampled as in-context examples, allowing the model to leverage higher-quality reasoning chains during refinement.

\paragraph{Iterative Refinement:} After initialization, we iteratively refine each article's reasoning chain (CoT) using in-context learning \cite{brown2020language}, thereby improving its predictions. In each refinement step for a given instance, we sample CoTs from  $C$ to construct K-shot in-context examples using the I-bias strategy. These examples are fed into the LLM along with the target article $a_i$ and instruction $\mathcal{I}$ to produce new refined reasoning chains and predictions conditioned on the examples provided. Illustrated as \begin{math}(\tilde{c}_i, \tilde{y}_i) \gets \text{LLM}( a_i, S_K, \mathcal{I})\end{math}, here $S_K$ is the sampled in-context examples of size K. If the new prediction achieves a higher $F1_{LCS}$ score compared to the previous version for a given instance $i$, we accept the updated pair $(\tilde{c}_i, \tilde{y}_i)$ and replace the old one, as shown in Figure \ref{fig: experiments2}. This process is repeated across the entire dataset and for multiple iterations.

\begin{figure}[t]
    \centering
    \includegraphics[width=0.8\linewidth]{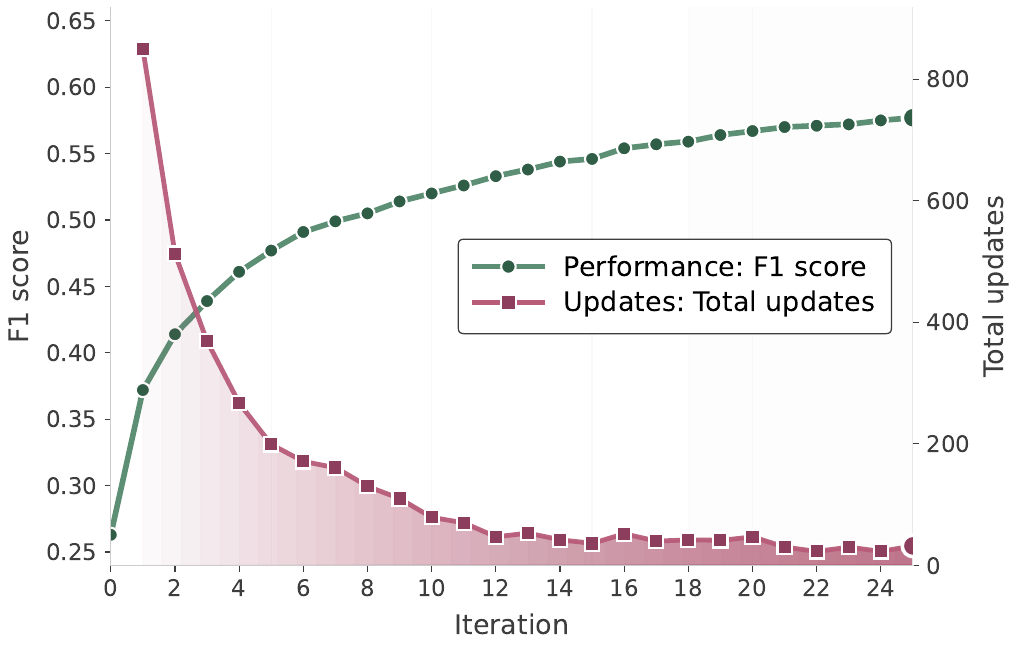}
    \caption{\model{'s} refinement and update density convergence curve across $25$ iterations for \textit{i-bias \& k=2} configuration on the \dataset{} dataset}
    \label{fig:refinement}
\end{figure}

\paragraph{Inference:} After refinement, $R$ examples are sampled from the evolved global pool and inserted into the prompt as few-shot examples, to produce predicted spans and reasoning for the test article, evaluated using the metric in \S\ref{sec:eval_mat}.

\section{Experiments, Results, and Analyses}

We use Mistral-Small-3.2-24B-Instruct for both initialization and refinement, for $n=25$ refinement iterations with $K \in \{1,2,3\}$ in-context example. The temperature is set to 1.0 during both refinement and inference. The prompts used are provided in Appendix~\ref{Appen_prompt}. We evaluate three inference backbones: Mistral-Small-3.2-24B-Instruct, Qwen3-32B, and GPT-4o-mini. To study the effect of the number of in-context examples during inference, we experiment with five setups where $R \in \{1,2,3,4,5\}$. The refinement and inference are conducted on the proposed \dataset{} and \PTC{} \footnote{A widely used benchmark for propaganda detection introduced in SemEval-2020 Task 11 for research and academic purposes.} \cite{da-san-martino-etal-2020-semeval}, and the reasoning pool is refined independently on both datasets. The average inference time per instance during inference is detailed in Appendix \ref{Appen_res}. We also fine-tuned Mistral-Small-3.2-24B-Instruct (SFT) as an upper-bound reference.

\input{New_Tables/results_main}

\paragraph{Baselines:}

We compared against \textbf{SemEval Models:} Aschern \cite{chernyavskiy-etal-2020-aschern} and Hitachi \cite{morio-etal-2020-hitachi-semeval},  \textbf{Prompting baselines}: Zero, One, Two-shot, and CoT \cite{wei2022chain} settings. \textbf{Reasoning baselines}: the RePrompting algorithm \cite{xu2023reprompting}, a population-based refinement baseline that employs Gibbs sampling, and EvoPrompt \cite{guo2023evoprompt}, based on evolutionary algorithms. and for a more recent baseline, we also included GEPA \cite{agrawal2026gepareflectivepromptevolution}

\paragraph{Evaluation Metrics:} \label{sec:eval_mat} 
We adopt evaluation metrics based on the longest common subsequence, given in equations \ref{eq:Plcs} and \ref{eq:Rlcs}, where $S_{pred}$ denotes predicted spans and $S_{gt}$ denotes ground truth spans. To compute $F1_{LCS}$ we take harmonic mean of $P_{LCS}$ and $R_{LCS}$. $F1_{LCS}$ evaluates both refinement and inference. 

\begin{equation} \small
P_{LCS} = \frac{1}{|S_{pred}|} \sum_{p \in S_{pred}} \max_{g \in S_{gt}} \left( \frac{LCS(p, g)}{|p|} \right) 
\label{eq:Plcs}
\end{equation}

\begin{equation} \small
R_{LCS} = \frac{1}{|S_{gt}|} \sum_{g \in S_{gt}} \max_{p \in S_{pred}} \left( \frac{LCS(p, g)}{|g|} \right)
\label{eq:Rlcs}
\end{equation}

\noindent Early iterations might result in a partially correct answer, an exact match criterion will reject the CoTs of early iterations, eliminating the learning cue, whereas $F1_{LCS}$  rewards partial token overlap, allowing the pool to progress gradually from a few correct tokens toward full span coverage across iterations.

\input{tables/i_bias_ablation}

\paragraph{Global Pool Convergence:}Figure \ref{fig:refinement} shows the refinement of the global reasoning pool $C$. The figure plots the relationship between update density and $F1_{LCS}$ score across 25 iterations on the Mistral backbone for the \dataset{} dataset (\PTC{} convergence curve is given in Appendix \ref{Appen_convergence}). The refinement curve exhibits a two-phase trajectory: an initial phase of rapid performance improvement followed by convergence in later iterations. In the early stages of refinement, the pool contains noisy reasoning chains. Consequently, this results in a large number of updates. Through population-level cross-example generalisation and I-bias strategy, \model{} progressively reinforces reasoning chains that capture useful cues. As refinement progresses, the performance gains gradually saturate, indicating that the reasoning pool has stabilised and that further updates provide only limited improvement. Consequently, the number of accepted updates also decreases in later iterations. 
Representative examples comparing zero-shot and refined CoTs are presented in Appendix \ref{Appen_examples} for the comparison of refined CoT with initialisation. The zero-shot CoT often relies on generic reasoning and extracts broader or less relevant spans. After iterative refinement, the CoTs become more focused on the defining characteristics of the target technique, resulting in more precise evidence selection.

\paragraph{Comparative Results: } 

Table \ref{tab:res} presents the quantitative analysis. 
Comparing with SFT (Upper bound reference), \model{} $R=3$ (Mistral) on \dataset{} achieves 88\% of the SFT upper bound on Avg $F1_{LCS}$ and 64\% on W.Avg $F1_{LCS}$. The larger gap on the weighted metric is consistent with SFT's advantage on high-frequency techniques, which benefit from direct gradient-based adaptation. Similarly, on \PTC{} $R=4$ (Qwen), \model{} reaches 85\% of the SFT upper bound on Avg $F1_{LCS}$ and 79\% on W.Avg $F1_{LCS}$, closing much of this gap while remaining interpretable. 

\dataset{} with Mistral and Qwen at R=3 achieves the best macro $F1_{LCS}$ of 0.364 and weighted $F1_{LCS}$ of 0.362, respectively. These results surpass all prompting and reasoning baselines. Mistral configuration improves macro $F1_{LCS}$ by +0.101, +0.046 and +0.038 over zero, one-shot and RePrompting, respectively.  On \PTC{}, \model{} delivers comparable rather than superior results relative to the baselines; the best \model's configuration (Qwen, $K=2$, $R=4$) achieves 0.200 average and 0.211 weighted $F1_{LCS}$. We hypothesise that, in the \PTC{} dataset, \model{} 's performance degradation is driven by domain/topic diversity in the in-context examples selected during inference. \dataset{} is drawn from two closely related, temporally concentrated events (CAA and the farmers' protest), so examples sampled from the reasoning pool at inference time share the same domain and framing, even though selection is random. \PTC{} spans many unrelated events, so random sampling at inference more often pulls in-context examples from a different topic/domain than the test instance. This shows context/domain dependence of \model{} (A detailed explanation is provided in Appendix \ref{Appen_convergence}). We illustrate this with representative cases below:

\begin{itemize}[leftmargin=*, nolistsep, noitemsep]
    \item \PTC{} - [Article: Glamorising (US Military) + R=3 in-context examples: Neutral reporting (Iran regime) + criticise(American President)+ Report(Plague in Madagascar), showing mismatched domains]   
    \item \dataset{} - [Article: criticise(CAA protest) + R=3 in-context examples: criticise(CAA protest) + criticise(Farmers' protest) + criticise(Govermnt over CAA bill), showing same-event/domain alignment]
\end{itemize}

\noindent Overall, R=3 \& R=4 perform better than other configurations, as a smaller $R$ provides insufficient in-context examples and a larger $R$ introduces more noise. $R_{LCS}$ and $P_{LCS}$ for K=2 configuration is reported in Appendix \ref{Appen_res} along with the quantitative results for K=1 and K=3 configurations.  We also evaluated using the Sem-Eval metrics introduced in \newcite{da2019fine}, the results for the Mistral backbone are given in Table \ref{tab:semeval_results}. Full table is provided in the Appendix\footnote{Due to the brevity, only the Mistral backbone is reported here in Table \ref{tab:semeval_results}.} \ref{Appen_res},  On \dataset{}, \model{} outperforms both SemEval models and standard prompting baselines on both backbones, peaking at avg. F1 0.277 (R=3) / w.F1 0.280 (R=2). On \PTC{}, scores are lower overall, with \model{} peaking at avg. F1 0.120 / w.F1 0.164 ($R=2$). These results mirror the trend observed with our LCS-based metric: \PTC{} scores are lower across all methods, including SFT, reflecting the benchmark's intrinsically difficult nature.

\input{New_Tables/semeval}

\input{tables/error-analysis}

\input{tables/humaneval}

\input{New_Tables/cross}

\paragraph{Ablation:}
We also report the effect of in-context examples during refinement, Table \ref{tab:i_bias_ablation} analyses the effect of varying in-context examples $K \in \{1,2,3\}$ on the refinement process. K=2 consistently provides the strongest overall performance across both datasets, particularly for configurations with R=3 and R=4. Whereas K=1 provides a limited example for cross-example generalisation, restricting the model's ability to accurately capture cues across persuasion techniques. In contrast, larger configurations, such as K=3, introduce noise, which negatively affects the refinement process. These results suggest that moderate population-level interaction during refinement is sufficient for capturing and technique-aware reasoning patterns. Whereas \ref{cross} reports cross-dataset transfer performance for the Mistral ($K=2$) configuration. On \PTC{}, cross-dataset transfer performs comparably to, and in fact exceeds, same-dataset inference: average F1 ranges from 0.185 to 0.224 across $R$, compared to 0.155--0.173 when \model{} is refined and evaluated on \PTC{} alone (Table \ref{tab:res}, $R=3$). On \dataset{}, cross-dataset transfer shows a modest drop across $R$: average F1 ranges from 0.286 to 0.327, compared to 0.320--0.351 when \model{} is both refined and evaluated on \dataset{} itself (Mistral backbone, $R=3$). These results suggest that CoT pools refined on one dataset transfer effectively to the other without requiring re-refinement, and in the case of \PTC{}, even yield improved performance likely because the \dataset{}-refined pool encodes more diverse, less domain-anchored reasoning cues that generalise better to \PTC{}'s heterogeneous topics. The same-dataset comparison figures are drawn from the best-performing $R=3$ configuration (Mistral) in Table \ref{tab:res}. In addition to the proposed selection strategy, we evaluate different bias strategies to examine the effects of exploratory and exploitative biases during sampling and in the refinement process. A detailed discussion of bias strategies is in Appendix \ref{Appen_Bias} and \ref{Appen_res}.

\FloatBarrier

\paragraph{Error Analysis:}
Error analysis reveals four failure modes: over-prediction, under-prediction, false positives, and false negatives, as shown in Table \ref{Tab: error}. Over-prediction: model captures excessive contextual text beyond the gold span, reducing precision. Under-prediction: arises from missing text, which impacts recall. False positives triggered by strong lexical cues (e.g., legal terms) in otherwise factual text. False negatives primarily involve subtle, implicit, or sarcastic propaganda, indicating difficulty in capturing nuanced pragmatic cues. These errors highlight the challenges of accurately detecting span boundaries.

\subsection{Human analysis}
To qualitatively assess the refined CoT explanations, we evaluate CoTs from iterations 0 and 25, 20 CoT instances were evaluated at iteration 0 and iteration 25, rated by 11 human evaluators and 9 expert evaluators. Human evaluators focused on interpretability-related factors. And experts trained in propaganda detection assessed CoTs for technical grounding, emphasising technique purity and logical structure. Each criterion defined for analysis is scored as poor, fair, and good, as described in Table \ref{tab: human_expert_eval}. To reduce evaluation bias, CoTs were anonymised, and Evaluators were not informed of the CoT's origin. 

\paragraph{Human Evaluation Criteria: } Assess the interpretability and readability of refined CoTs, particularly how well the reasoning justifies the extracted information from the source text.
    \textbf{Clarity of Reasoning(CR)}: Measures whether the CoT is clear and whether each step is easy to understand.
    \textbf{Relevance to the Text(RT)}: Assesses whether the reasoning refers only to the article content and avoids hallucinations. It also judges whether the spans extracted by the CoT are actually present and semantically meaningful.
    \textbf{Overall Coherence(OC)}: Rates whether the reasoning flows logically and is consistent with the final extracted spans.

\paragraph{Expert Evaluation Criteria:} The expert evaluators are the same annotators who contributed to the dataset and developed expertise in identifying propaganda spans with logical justification. They evaluate the logical structure and technique purity of the generated reasoning. \textbf{Technique Purity (TP)}: Determines whether the CoT adheres to the intended propaganda technique (e.g., Name-Calling/Labelling) without mixing unrelated techniques such as Appeal to Fear or Loaded Language. It also evaluates whether the CoT correctly explains and applies the technique's definition. \textbf{Logical Structure (LS)}: Assesses whether the reasoning is stepwise, methodical, and generalizable.

\section{Conclusion}

We present \model{}, an interpretable propaganda detection framework that provides explanations for identifying propagandistic spans, unlike standard black-box approaches. \model{} leverages population-level CoTs refinement, using a sampling strategy that prioritises CoTs that yield measurable performance gains and progressively captures reasoning cues. Along with interpretability, \model{} reaches up to 88\% and 85\% of a supervised fine-tuning (SFT) upper bound on Avg $F1_{LCS}$ on \dataset{} and \PTC{} respectively. With competitive performance across all baselines on the \dataset{} dataset, while \PTC{} shows modest performance; lower scores across all methods, including SFT, reflect the benchmark's intrinsically harder and more topic-diverse nature of \PTC{}.

\FloatBarrier
\section*{Limitations}

While \model{} shows promising results, it has several limitations. The homogeneous domain/event yields better results compared to heterogeneous domains. And the iterative refinement framework is computationally intensive, as it involves multiple refinement steps. Future work should explore lightweight refinement strategies, prompt compression, or smaller specialised models to improve efficiency. Errors or inaccuracies in text generated by large language models (LLMs) may affect the quality of the resulting reasoning chains. Currently, the framework is limited to English-language news articles; further studies can incorporate multilingual or code-mixed content.

\section*{Ethical Considerations}
This work aims to promote transparency in media analysis by helping researchers and readers to detect persuasive and potentially manipulative communication patterns in news articles. But it is necessary to deploy propaganda-detection technologies responsibly and to read them with caution. Automated propaganda detection systems are not to be used to determine the truthfulness, legitimacy, or political validity of opinions expressed in the media. The proposed framework detects persuasive patterns and produces reasoning-based explanations, but these outputs are intended to support human analysis rather than to substitute for human judgment. Users might be vulnerable to over-trusting automatically generated explanations by large language models. As hallucinations and factually inconsistent reasoning are known issues with LLMs, the system's explanations may sometimes contain inaccuracies or incomplete interpretations. So, human oversight and expert verification remain essential for interpreting the outputs. Propaganda detection systems might unintentionally reinforce societal or political biases embedded in the underlying datasets or language models. Therefore, such systems should be carefully evaluated in different domains, perspectives and demographic contexts before they are deployed in sensitive real-world applications. Finally, this research is not intended to promote censorship or restrict political speech. Instead, the proposed framework is intended to be an assistive analytical tool that fosters critical thinking and media literacy by helping users explore rhetorical and persuasive strategies in news discourse.

\bibliography{custom}

\appendix

\newpage

\FloatBarrier
\section{Bias Strategies}\label{Appen_Bias}

\subsection{Random (No Bias)} \label{sec: no_bias}

We define sampling from the list $C$ as follows. At each iteration, we aim to refine the CoT at a current index $i \in \{1, \dots, M\}$, which is excluded from the candidate pool to avoid self-conditioning. 

Let \( V \subseteq C \setminus \{i\} \) denote the set of valid indices corresponding to non-empty CoTs, excluding the index $i$. The sampling function $f(i, K)$ samples a subset with replacement, and each reasoning chain contains equal probability of selection.  \( S \subseteq V \) of size $K$, such that \( |S| = K \), Where $K$ is the number of in-context support examples. The formal definition is given in equation \ref{eq:rnd}. 

\begin{equation}\small
f(i, K) =
\begin{cases}
\text{Sample}(V, K), & \text{if } |V| \geq K \\
\emptyset, & \text{if } |V| < K
\end{cases}
\label{eq:rnd}
\end{equation}

The sampling strategy ensures that each refinement step is based on a randomly selected set of CoTs and their corresponding prediction.

\subsection{Sampling with Bias on Updating Index} \label{sec: j_bias}
\label{sec:sample_updating_bias}
We define a biased sampling strategy that favors CoTs which have previously contributed to the improvement of other reasoning chains. At each iteration, we aim to refine the CoT at a current index $i \in \{1, \dots, M\}$, which is excluded from the candidate pool to avoid self-conditioning.

Let $V \subseteq C \setminus \{i\}$ denote the set of valid indices corresponding to non-empty CoTs, excluding index $i$. We maintain a global probability array $\mathbf{p} \in \mathbb{R}^{M}$, initialized uniformly as:

\begin{equation}\small
p_j^{(0)} = \frac{1}{M}, \quad \forall j \in \{1, \dots, M\}
\label{eq:uniform_init_updating}
\end{equation}

After each refinement step, if the CoT at index $i$ yields a higher F1 score than its previous version, the probability of the support index $j \in S$ ($S$ as defined in section \ref{sec: no_bias}) that caused the update is incremented by a fixed bias $\delta = 0.02$, subject to a maximum cap of $p_{\max} = 0.1$:

\begin{equation}\small
\hat{p}_j^{(t)} = \min\left(p_j^{(t-1)} + \delta,\ p_{\max}\right)
\label{eq:bias_update_updating}
\end{equation}

The global probability array is then immediately renormalized to maintain a valid distribution:

\begin{equation}\small
p_k^{(t)} = \frac{\hat{p}_k^{(t)}}{\sum_{l=1}^{M} \hat{p}_l^{(t)}}, \quad \forall k \in \{1, \dots, M\}
\label{eq:renorm_updating}
\end{equation}

where $\hat{p}_k^{(t)} = p_k^{(t-1)}$ for all $k \neq j$, i.e., only the support index $j$ is modified before renormalization.

The sampling function $f(i, K)$ then draws $K$ support indices from $V$ using the restricted and renormalized probability distribution over candidates:

\begin{equation}\small
f(i, K) =
\begin{cases}
\text{Sample}\left(V,\ K,\ \mathbf{p}_V^{(t)}\right), & \text{if } |V| \geq K \\
\emptyset, & \text{if } |V| < K
\end{cases}
\label{eq:updating_bias_sampling}
\end{equation}

where $\mathbf{p}_V^{(t)}$ is the renormalized distribution restricted to $V$. This strategy ensures that CoTs which have consistently acted as useful support examples for improving other reasoning chains are more frequently sampled, allowing the model to prioritize high-quality and influential reasoning structures during refinement.

\section{Dataset Statistics} \label{Appen_Stats}

The dataset is annotated for persuasive spans for 20 techniques. The corpus contains 3922 annotated spans, out of which 3126 are allocated to the train set and 796 to the test set.

\subsection{Technique-wise Distribution}

The dataset identifies 20 specific propaganda techniques. The distribution across these techniques is characterized by significant variance, reflecting the real-world prevalence of certain rhetorical strategies over others.

\begin{figure}[t]
    \centering
    \includegraphics[width=1\linewidth]{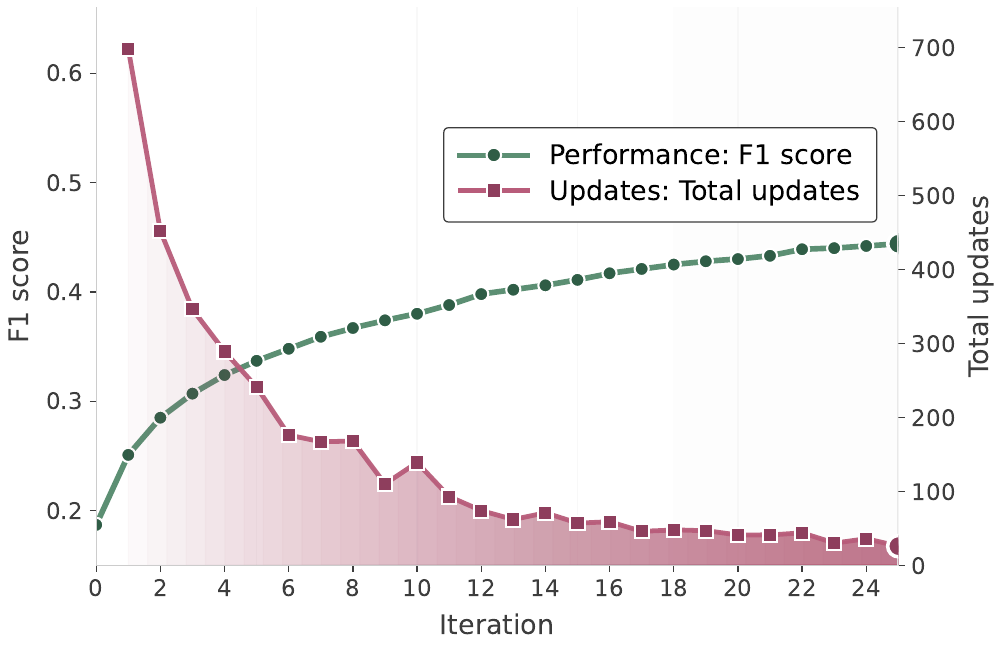}
    \caption{\model{'s} refinement and update density convergence curve across $25$ iterations for \textit{i-bias \& k=2} configuration on the \PTC{} dataset}
    \label{fig:ptc_update}
\end{figure}

\paragraph{High-Frequency Techniques}
The most prevalent technique across both datasets is Smears (T15), totaling 851 spans. This is followed by Assertion (T19, 561 spans) and Appeal to emotion/fear/prejudice (T9, 515 spans). The prominence of these techniques indicates a heavy reliance on ad hominem attacks and emotional manipulation within the sampled discourse.

\paragraph{Low-Frequency Techniques}
 Conversely, techniques such as Reductio ad Hitlerum (T5) and Black and White Fallacy (T2) show significantly lower frequencies (28 and 32 total spans, respectively). 

\section{Prompt} \label{Appen_prompt}

\begin{figure*}[t]
    \centering
    \includegraphics[width=1\linewidth]{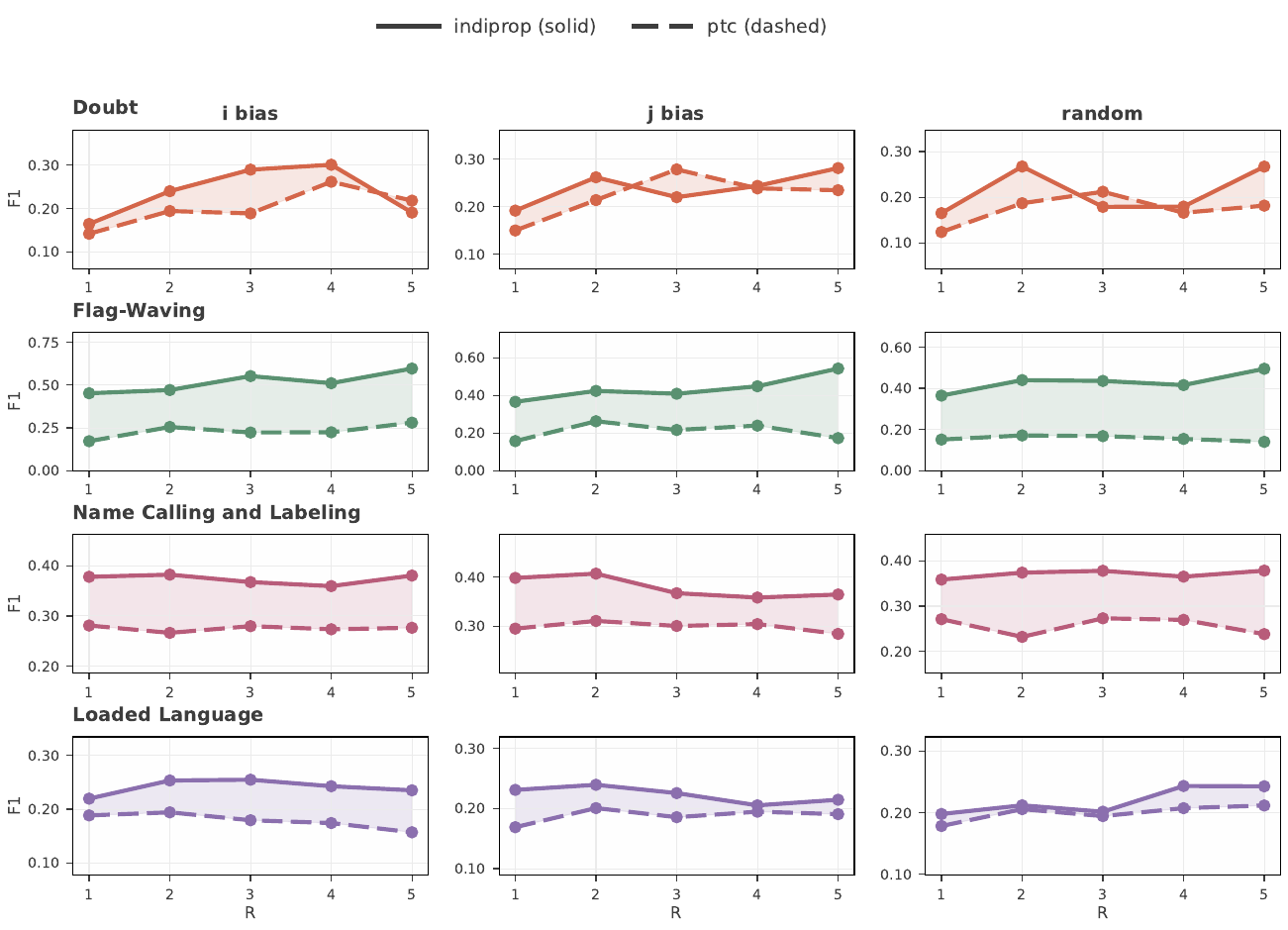}
    \caption{Inter-bias comparison of technique wise inference results across $R$ values for $K=2$ on Mistral backbone}
    \label{fig:inter_bias}
\end{figure*}

During iterative refinement, we employed a fixed Chain-of-Thought prompt that requires step-by-step reasoning for a target propaganda technique before any span extraction. Spans are produced only upon a positive reasoning conclusion based on $F_{LCS}$ scores, ensuring interpretable and evaluation-consistent refinement.
The prompts used during refinement and inference are included in Figure \ref{fig:prompts_combined}.

\input{New_Tables/inference_time}

\section{Global pool convergence over PTC}\label{Appen_convergence}

Figures \ref{fig:ptc_update} illustrate Global pool convergence curves over the \PTC{} dataset. We observe a similar convergence trend for \PTC{} as observed for \dataset{}, with a slight difference during the initial phase. While \dataset{} showed a steep performance increase early in the refinement, \PTC{} shows a more gradual increase through the initial phase. The update density convergence trend is similar to the \dataset{} dataset, with large updates in the initial phase, then steady near zero degrees by the final iteration. \dataset{} is an event-specific dataset, allowing in-context examples from the same event to yield rapid performance gains, whereas \PTC{} is not event-specific, leading to a steadier convergence trajectory. Overall, both curves corresponding to both datasets demonstrate convergence. 

\section{Inter-Bias Inference analysis}\label{Appen_bias_inference}

Figure \ref{fig:inter_bias} compares the techniques across all three bias strategies, showing that persuasive techniques exhibit the same trends for in-context example size during inference, indicating similar intrinsic cues acquired during the refinement process, with sampling strategy being i-bias. However, for j and random bias, this trend shows slight variations across frequent and infrequent techniques, such as doubt and flag-waving, which are less dominant. The trajectory of $F_{LCS}$ vs $R$ shows a difference, indicating that the refinement curve flattens but not towards meaningful intrinsic cues, as j-bias prioritises the CoTs that cause updates, resulting in an echo chamber of cues that were successful in earlier refinements but failed at inference. Similarly, random, as a stochastic process, was unable to capture meaningful reasons either. whereas dominant techniques, name-calling, and loaded language performed slightly better during refinement, due to the large population size. 

\section{Result Analysis}\label{Appen_res}

Table \ref{tab:k1_k3_ablation} reports of $R_{LCS}$, $P_{LCS}$, and $F1_{LCS}$ across all baselines and ReCoT configurations at K=1, K=2 and K=3 on both \dataset{} and \PTC{}, extending the main K=2 results in Table \ref{tab:res}.

 Mistral at K=1, R=3 achieves $F1_{LCS}$=0.354  and 0.360 for macro and weighted respectively and $R_{LCS}$=0.447 a substantial recall gain over zero-shot ($R_{LCS}$=0.270) and 1-shot ($R_{LCS}$=0.365), indicating that even a single in-context example from the refined pool is sufficient to improve span coverage. Qwen at K=1, R=3 achieves $F1_{LCS}$=0.321 and the highest weighted $P_{LCS}$ of 0.380 among K=1 configurations, suggesting that recot at this setup accurately captures span boundaries. GPT-4o-mini at K=1 shows the weakest performance across all R values, achieving the highest $F1_{LCS}$=0.307 at R=4 for macro as well as weighted and degrades at R=5, still surpassing its own zero-shot baseline (0.294) and CoT baseline (0.302), confirming that population-level refinement provides a consistent benefit regardless of backbone capability.

On \PTC{} dataset at K=1, Mistral achieves its best result at R=4 acheives $F1_{LCS}$=0.193 and 0.205 for macro and weighted, respectively, with Qwen performing comparably at R=1 and R=2 ($F1_{LCS}$ for W.Avg =0.201 - 0.205), both surpassing the Mistral prompting baselines on \PTC. At K=3, performance degrades relative to K=1 and K=2 configurations for all LLM backbones, with the weakest observed performance is Qwen, where $F1_{LCS}$ drops to 0.194 to as low as 0.166 on \PTC{} dataset, falling below zero-shot performance, this degradation at K=3 is also observed for macro recall and precision on \dataset{}, Mistral at K=3 maintains relatively high recall at R=5 ($R_{LCS}$=0.437-0.478) but shows markedly reduced precision ($P_{LCS}$=0.308-0.331) compared to baselines, indicating over-conditioning due to the presence of three in-context examples causes the model to over-predict spans, expanding coverage at the cost of boundary precision. 

For overall recall performance, GPT EvoPrompt achieves the highest recall on the \dataset{} dataset among all methods. But for the same precision, it is as low as 0.237-0.262, the lowest among all GPT configurations. The $F1_{LCS}$ collapses to 0.272-0.300. This behaviour indicates EvoPrompt's optimisation drives the model toward over-prediction, generating excessively long or numerous spans that happen to overlap with ground truth. Similar behaviour is noticed on the \PTC{} dataset, where recall reaches to 0.419 Avg and 0.366 W.Avg, but precision drops to 0.200-0.178. again indicating over-prediction. 

The performance of \model{} at K=2 in terms of precision and recall. Mistral at R=3 achieves the best \model{} $R_{LCS}$ of 0.457 for macro Avg and 0.446 for Weighted Avg. on \dataset{} dataset. Beyond R=3, recall drops slightly (R=4: 0.420, R=5: 0.429), indicating that R=3 strikes a balance between constrained and noisy contexts. Qwen achieves best macro $P_{LCS}$ at R=5: 0.368 and weighted $P_{LCS}$ at R=3: 0.387.  This behaviour indicates that Mistral's refined CoTs are coverage-oriented, whereas Qwen's refined CoTs are more boundary-precise, making fewer but more accurate predictions. Both strategies eventually converge to competitive F1 along different paths, indicating that the global pool converges in the end. EvoPrompt(GPT) achieved 0.419 macro $R_{LCS}$, 0.366 weighted $R_{LCS}$ on the \PTC{} dataset. As stated earlier, these are the result of overprediction. And a similar trend is observed for EvoPrompt on \dataset{}. 

Qwen achieves the best \model{} recall and the best \model{} precision on \PTC{} dataset simultaneously at R=4: macro $R_{LCS}$ of 0.256 and weighted macro $R_{LCS}$  of 0.250, which is why it also produces the best $F1_{LCS}$ of 0.200-0.211 (mentioned in Table \ref{tab:res}). Whereas GPT achieves macro recall of 0.342--0.362 on \dataset{} dataset, comparable to Qwen, but its weighted recall (0.225--0.248) is dramatically lower.
Overall, these results confirm that \model{} recall improvements are grounded in technique understanding rather than span inflation and achieve competitive precision alongside recall gains across backbones and datasets. This case provides evidence that population-level CoT refinement captures the intrinsic definition characteristic of the techniques 

\input{New_Tables/semeval_gpt}
\FloatBarrier

\paragraph{Inference Time:}
The average inference time per instance for the $R=3$ configuration for \dataset{} during infernce is approximately 6.05s for Mistral and 11.70s for Qwen. And, for the refinement phase, the runtime for a single instance within a single iteration is approximately 2.99s (Mistral) and 5.63s (Qwen) for K=1, 4.7s (Mistral) and 9.01s (Qwen) for K=2, and 6.12s (Mistral) and 12.0s (Qwen) for K=3. (as shown in table \ref{infer})

\paragraph{Technique Wise Performance:}

Tables~\ref{indi_tech} and~\ref{ptc_tech} report the per-technique F1 scores for the \textit{Mistral K2 -- iBias} configuration on the \dataset{} and \PTC{} datasets, respectively, when varying the number of in-context examples used during inference. Here, R1--R5 denote inference with one to five retrieved Chain-of-Thought (CoT) examples from the refined example pool for the corresponding propaganda technique. While the main paper reports aggregate performance, these tables provide a fine-grained view of how increasing the amount of retrieved in-context evidence affects individual techniques.

\input{New_Tables/tech_wise_indi}

Across both datasets, the influence of additional in-context examples is technique-dependent rather than monotonic. Some techniques, such as \textit{Flag Waving}, benefit from additional examples, whereas others, including \textit{Loaded Language} and \textit{Name Calling and Labelling}, remain comparatively stable. Moreover, the same technique does not necessarily exhibit similar trends across datasets; for instance, \textit{Bandwagon} shows contrasting behaviour on IndiProp and PTC. These observations suggest that the optimal amount of retrieved CoT evidence depends on both the propaganda technique and the dataset, reflecting differences in class distributions and the linguistic characteristics of the two corpora.

\paragraph{Bias strategies: } In addition to the proposed selection strategy, we evaluate different bias strategies to examine the effects of exploratory and exploitative biases during sampling and in the refinement process. \textbf{Random sampling:} This sampling strategy is randomised, that is, it uniformly samples without any preference from the valid set of indices explained in Section \ref{sec: i_bias}. \textbf{J-bias:} It increases the probability of CoTs that have influenced and improved other reasoning chains. Table \ref{tab: ablation1} compares I-Bias, Random, and J-Bias across $R \in \{1,2,3,4,5\}$ with K=2 on Mistral and Qwen inference backbones. Random at R=5 achieved the highest macro $F1_{LCS}$ of 0.378 on Mistral backbone for the \dataset{} dataset, but the same configuration yields the lowest weighted $F1_{LCS}$ of 0.111 on \PTC. As random is a stochastic process, consistent behaviour cannot be guaranteed. J-Bias, exclusively prioritising CoTs that have previously updated other CoTs, risks creating an echo chamber of dominant reasoning cues that were successful in the early refinement phase. I-Bias, by contrast, balances exploration and exploitation by rewarding CoTs based on their own performance improvement while keeping the bias increment small to retain diversity.

\input{New_Tables/tab_ablation}

\input{New_Tables/tech_wise_ptc}

\paragraph{SemEval Span-overlap Metrics: }
 We reran the evaluation using SemEval's official span-overlap metrics, as shown in Table \ref{tab:gpt_semeval_results}. On Indiprop, ReCoT outperforms both SFT and standard prompting baselines on both backbones, peaking at avg. F1 0.272 / w.F1 0.274 (GPT R5). On PTC, scores are lower overall but follow the same trend, with GPT ReCoT peaking at R4 (avg. F1 0.130, w.F1 0.161), close to the strongest prompting baseline (GPT 1-shot, w.F1 0.203). These results mirror the trend we reported with our LCS-based metric.

\section{Before-and-After examples}\label{Appen_examples}

Table \ref{tab: propaganda_spans} compared iteration 0 with iteration 25 across all bias strategies. Compared to the zero-shot initialization, all refined variants produce more focused reasoning. I-Bias learns to directly evaluate whether the article misrepresents the opponent's viewpoint, resulting in a concise CoT that identifies the key propagandistic span and achieves the highest F1 (0.706). J-Bias improves over initialization by providing a more structured explanation that better captures multiple propaganda cues, although its broader reasoning extracts an additional, less relevant span, leading to a lower F1 than I-Bias. Random refinement is also more targeted than the initial zero-shot CoT, but its inconsistent exploration identifies only a single partial span, yielding only a modest improvement over initialization. Overall, iterative refinement makes the reasoning progressively more focused and evidence-driven compared to the unrefined zero-shot CoT.

\FloatBarrier

\input{tables/k1_ablation}
\input{tables/cot_compare}

\begin{figure*}[t]
\centering

\begin{tcolorbox}[
  colback=white,
  colframe=black,
  coltitle=white,
  colbacktitle=gray,
  title={Prompt Used During Iterative Refinement},
  fonttitle=\bfseries,
  sharp corners,
  boxrule=1pt,
  width=0.95\textwidth,
]
\footnotesize
You are an expert propaganda span detector.  
Your task is to carefully analyze the article and decide step by step
(Chain of Thought) whether there are any statements that qualify as
propaganda of type \texttt{\{technique\}}.

\vspace{0.5em}
\textbf{Instructions:}
\begin{enumerate}
    \item Use a step-by-step reasoning process inside \texttt{<CoT>...</CoT>} tags.
    \item If no spans are found, clearly state that in the CoT and output nothing inside \texttt{<span>} tags.
    \item If spans are found, wrap each exact substring from the article in \texttt{<span>...</span>} tags.
\end{enumerate}

\vspace{0.5em}
\textbf{Output format (exactly):}
\begin{verbatim}
<CoT>
   Step 1: ...
   Step 2: ...
   Step 3: ...
   Conclusion: Propaganda spans exist / do not exist
</CoT>

<!-- zero or more spans, depending on conclusion -->
<span>...</span>
<span>...</span>
\end{verbatim}
\end{tcolorbox}

\vspace{0.8em}

\begin{tcolorbox}[
  colback=white,
  colframe=black,
  coltitle=white,
  colbacktitle=gray,
  title={Prompt Used During Inference},
  fonttitle=\bfseries,
  sharp corners,
  boxrule=1pt,
  width=0.95\textwidth,
]
\footnotesize
Here is an example: \{support text\}

Target Article: \{full article\}

You are an expert propaganda span detector.  
Your task is to carefully analyze the article and decide step by step
(Chain of Thought) whether there are any statements that qualify as
propaganda of type \texttt{\{technique\}}.

\vspace{0.5em}
\textbf{Instructions:}
\begin{enumerate}
    \item Use a step-by-step reasoning process inside \texttt{<CoT>...</CoT>} tags to decide if such propaganda spans exist.
    \item If no spans are found, clearly state that in the CoT and output nothing inside \texttt{<span>} tags.
    \item If spans are found, wrap each exact substring from the article in \texttt{<span>...</span>} tags.
\end{enumerate}

\vspace{0.5em}
\textbf{Output format (exactly):}
\begin{verbatim}
<CoT>
   Step 1: ...
   Step 2: ...
   Step 3: ...
   Conclusion: Propaganda spans exist/do not exist
</CoT>

<!-- zero or more spans, depending on conclusion -->
<span>...</span>
<span>...</span>
\end{verbatim}
\end{tcolorbox}

\caption{Prompts used in our framework: (top) iterative refinement prompt, (bottom) inference prompt.}
\label{fig:prompts_combined}

\end{figure*}

\end{document}

%% file: tables/technique_frequency.tex
\begin{table}[t]
\centering
\tiny
\setlength{\tabcolsep}{3pt}
\renewcommand{\arraystretch}{1.1}
\begin{tabular}{lp{3.6cm}cccc}

\toprule
ID & Technique & Train & Test & Total \\
\midrule
T1  & Causal Oversimplification & 99  & 21 & 120 \\
T2  & Black-and-white Fallacy & 28 & 4 & 32 \\
T3  & Straw Man & 55 & 17 & 72 \\
T4  & Whataboutism & 32 & 10 & 42 \\
T5  & Reductio ad Hitlerum & 22 & 6 & 28 \\
T6  & Red Herring & 31 & 13 & 44 \\
T7  & Loaded Language & 273 & 69 & 342 \\
T8  & Name Calling and Labeling & 235 & 67 & 302 \\
T9  & Appeal to emotion/fear/prejudice & 411 & 104 & 515 \\
T10 & Slogans & 164 & 41 & 205 \\
T11 & Flag-Waving & 77 & 21 & 98 \\
T12 & Exaggeration and Minimization & 56 & 12 & 68 \\
T13 & Thought-terminating Clich\'es & 85 & 18 & 103 \\
T14 & Bandwagon & 44 & 8 & 52 \\
T15 & Smears & 671 & 180 & 851 \\
T16 & Obfuscation, Vagueness, Confusion & 29 & 6 & 35 \\
T17 & Doubt & 124 & 25 & 149 \\
T18 & Appeal to Authority & 58 & 14 & 72 \\
T19 & Assertion & 449 & 112 & 561 \\
T20 & Glittering Generalities (Virtue) & 100 & 28 & 128 \\
\midrule
\textbf{Total} & \textbf{All Techniques} & \textbf{3126} & \textbf{796} & \textbf{3922} \\
\bottomrule

\end{tabular}
\caption{Frequency distribution of propaganda techniques in \dataset{}.}
\label{tab:dataset_stats_compact}
\end{table}

%% file: New_Tables/results_main.tex
\begin{table}[t]
\centering
\scriptsize
\setlength{\tabcolsep}{2pt}
\renewcommand{\arraystretch}{0.85}
\begin{tabular}{l l cc cc}
\toprule
\multirow{2}{*}{\textbf{Config}}
& \multirow{2}{*}{\textbf{Backbone}}
& \multicolumn{2}{c}{\textbf{IndiProp}}
& \multicolumn{2}{c}{\textbf{PTC}} \\
\cmidrule(lr){3-4}
\cmidrule(lr){5-6}
& & Avg & W.Avg & Avg & W.Avg \\
\midrule
SFT (Upper Bound)  &   Mistral                       & 0.412	& 0.538 &	0.235 &	0.266 \\
\midrule
Hitachi  & \multirow{2}{*}{---} & 0.180 & 0.232 & 0.093 & 0.118 \\
Aschern  &                      & 0.205 & 0.231 & 0.174 & 0.203 \\
\midrule
0-shot & \multirow{4}{*}{Mistral} & 0.263 & 0.273 & 0.187 & 0.184 \\
1-shot &                          & 0.318 & 0.317 & 0.200 & 0.192 \\
2-shot &                          & 0.154 & 0.157 & 0.195 & 0.194 \\
CoT    &                          & 0.285 & 0.300 & 0.186 & 0.202 \\
\midrule
RePrompt  &\multirow{2}{*}{Mistral}    & 0.326 & 0.314 & 0.160 & 0.186 \\
EvoPrompt &     & 0.213 & 0.227 & 0.171 & 0.195 \\
GEPA  &                          & 0.296 & 0.324 & \textbf{0.233} & 0.278 \\
\midrule
0-shot & \multirow{4}{*}{GPT} & 0.294 & 0.279 & 0.169 & 0.163 \\
1-shot &                      & 0.254 & 0.263 & 0.218 & \textbf{0.215} \\
2-shot &                      & 0.275 & 0.279 & 0.212 & 0.203 \\
CoT    &                      & 0.302 & 0.289 & 0.210 & 0.208 \\
\midrule
RePrompt  & \multirow{2}{*}{GPT} & 0.297 & 0.308 & 0.193 & 0.209 \\
EvoPrompt &                     & 0.272 & 0.300 & 0.232 & 0.208 \\
\midrule
\model{} R3 &    Mistral                       & \textbf{0.364} & 0.347 & 0.155 & 0.172 \\

\midrule

\model{} R3 &     \multirow{2}{*}{Qwen}                & 0.345 & \textbf{0.362} & 0.184 & 0.198 \\
\model{} R4 &                       & 0.326 & 0.356 & 0.200 & 0.211 \\

\midrule

\model{} R3 &            GPT          & 0.320 & 0.216 & 0.165 & 0.130 \\
\bottomrule
\end{tabular}
\caption{$F1_{LCS}$ Performance comparison on \dataset{} and \PTC{} for \textit{i-bias \& K=2 setup}.}
\label{tab:res}
\end{table}

%% file: tables/i_bias_ablation.tex
\begin{table}[t]
\centering
\resizebox{\columnwidth}{!}{
\begin{tabular}{llcccccccc}
\toprule
\multirow{2}{*}{Setup} & \multirow{2}{*}{R}
& \multicolumn{4}{c}{Mistral}
& \multicolumn{4}{c}{Qwen} \\
\cmidrule(lr){3-6} \cmidrule(lr){7-10}
&
& \multicolumn{2}{c}{IndiProp}
& \multicolumn{2}{c}{PTC}
& \multicolumn{2}{c}{IndiProp}
& \multicolumn{2}{c}{PTC} \\
\cmidrule(lr){3-4} \cmidrule(lr){5-6} \cmidrule(lr){7-8} \cmidrule(lr){9-10}
&
& Avg & W.Avg & Avg & W.Avg
& Avg & W.Avg & Avg & W.Avg \\
\midrule

\multirow{5}{*}{$K=1$}
& R=1 & 0.323 & 0.336 & 0.171 & 0.185 & 0.323 & 0.339 & 0.184 & 0.201 \\
& R=2 & 0.340 & 0.337 & 0.171 & 0.185 & 0.313 & 0.344 & 0.190 & 0.205 \\
& R=3 & 0.354 & \textbf{0.360} & 0.182 & 0.196 & 0.321 & 0.358 & 0.187 & 0.199 \\
& R=4 & 0.334 & 0.345 & \textbf{0.193} & \textbf{0.205} & 0.341 & 0.352 & 0.177 & 0.192 \\
& R=5 & 0.357 & 0.360 & 0.186 & 0.196 & 0.323 & 0.342 & 0.170 & 0.185 \\

\midrule
\multirow{5}{*}{$K=2$}
& R=1 & 0.320 & 0.311 & 0.167 & 0.180 & 0.296 & 0.326 & 0.172 & 0.191 \\
& R=2 & 0.343 & 0.300 & 0.166 & 0.182 & 0.316 & 0.348 & 0.181 & 0.196 \\
& R=3 & \textbf{0.364} & 0.347 & 0.155 & 0.172 & 0.345 & \textbf{0.362} & 0.184 & 0.198 \\
& R=4 & 0.344 & 0.317 & 0.172 & 0.170 & 0.326 & 0.356 & \textbf{0.200} & \textbf{0.211} \\
& R=5 & 0.361 & 0.306 & 0.172 & 0.170 & \textbf{0.359} & 0.355 & 0.170 & 0.187 \\

\midrule
\multirow{5}{*}{$K=3$}
& R=1 & 0.347 & 0.333 & 0.156 & 0.153 & 0.330 & 0.343 & 0.194 & 0.204 \\
& R=2 & 0.319 & 0.340 & 0.157 & 0.157 & 0.310 & 0.343 & 0.183 & 0.199 \\
& R=3 & 0.334 & 0.338 & 0.168 & 0.165 & 0.340 & 0.355 & 0.178 & 0.192 \\
& R=4 & 0.324 & 0.331 & 0.153 & 0.152 & 0.326 & 0.343 & 0.166 & 0.182 \\
& R=5 & 0.328 & 0.347 & 0.156 & 0.156 & 0.343 & 0.353 & 0.180 & 0.183 \\

\bottomrule
\end{tabular}
}
\caption{\model{} Performance across different values of $K$ using $i_{bias}$ strategy}
\label{tab:i_bias_ablation}
\end{table}

%% file: New_Tables/semeval.tex
\begin{table}[ht]
\centering
\scriptsize
\setlength{\tabcolsep}{5pt}
\renewcommand{\arraystretch}{0.9}
\begin{tabular}{l l cc cc}
\toprule
\multirow{2}{*}{\textbf{Config}} &
\multirow{2}{*}{\textbf{Backbone}} &
\multicolumn{2}{c}{\textbf{IndiProp}} &
\multicolumn{2}{c}{\textbf{PTC}} \\
\cmidrule(lr){3-4} \cmidrule(lr){5-6}
& & Avg & W.Avg & Avg & W.Avg \\
\midrule
SFT (Upper Bound)& Mistral & 0.210	& 0.214   & 0.127	 & 0.195  \\
\midrule
Hitachi & \multirow{2}{*}{---} & 0.161 & 0.207 & 0.092 & 0.143 \\
Aschern & & 0.152 & 0.193 & 0.089 & 0.141 \\
GEPA & Mistral & 0.247  &	0.256	& 0.158	   &  \textbf{0.185}\\
\midrule
0-shot & \multirow{3}{*}{Mistral} & 0.201 & 0.213 & 0.111 & 0.139 \\
1-shot & & 0.244 & 0.242 & \textbf{0.159} & 0.158 \\
CoT & & 0.213 & 0.235 & 0.109 & 0.148 \\
\midrule
RePrompt & \multirow{6}{*}{Mistral} & 0.263 & 0.259 & 0.147 & 0.168 \\
\model{} R1 & & 0.222 & 0.252 & 0.098 & 0.144 \\
\model{} R2 & & 0.272 & \textbf{0.280} & 0.120 & 0.164 \\
\model{} R3 & & \textbf{0.277} & 0.276 & 0.093 & 0.151 \\
\model{} R4 & & 0.258 & 0.258 & 0.106 & 0.155 \\
\model{} R5 & & 0.268 & 0.277 & 0.103 & 0.135 \\
\bottomrule
\end{tabular}
\caption{Performance comparison on the SemEval metrics - Avg F1 and Weighted Average (W.Avg) F1 are reported for the IndiProp and PTC tasks.}
\label{tab:semeval_results}
\end{table}

%% file: tables/error-analysis.tex
\begin{table}[t]
\centering
\renewcommand{\arraystretch}{1.3}
\resizebox{\columnwidth}{!}{
\begin{tabular}{p{7.5cm}p{6.5cm}}
\toprule
\textbf{Category of error} & \textbf{Remark / Error Analysis} \\
\midrule  

\textbf{GT: }[ORG]\footnote{The names of the organisation and policy are intentionally omitted to avoid explicit political content in the paper.} was misleading people by saying that [POLICY] is against minorities \newline
\textbf{Pred: }[ORG] was misleading people by saying that [POLICY] is against minorities, \textcolor{red!70}{and asked them to show proof for that claim }\textcolor{blue!70}{\textbf{[Over-prediction]}}

& Model added extra explanatory clause (``asked them to show proof''), extending beyond the gold boundary, another over-prediction error. \\

\midrule

\textbf{GT: }\textcolor{red!80}{The [ORG] is} targeting a particular community \textcolor{red!80}{by carrying out the [POLICY] in North and South 24 Pargana districts }\textcolor{blue!80}{\textbf{[Under-prediction]}}
\newline 
\textbf{Pred: }targeting a particular community 

& Model extracted only the core phrase, ignoring key qualifiers (``[ORG]'', ``[POLICY]''), leading to a truncated reasoning span. \\

\midrule

\textbf{GT: }No Propaganda span present. \newline
\textbf{Pred: } \textcolor{red!80}{what they did was `unlawful' and `not legally correct'}\textcolor{blue!80}{\textbf{[False Positive]}}

& Model predicted a span even though no persuasive content was annotated. False trigger due to lexical cues like ``unlawful'' and ``not legally correct''. \\

\midrule

\textbf{GT: }\textcolor{red!80}{They won't be able to give citizenship to anyone in two days. This is just a lollipop and a show-off.}\textcolor{blue!80}{\textbf{[False Negative]}} \newline
\textbf{Pred: }No Propaganda span present.

& Model failed to detect explicit ridicule and exaggeration, missing emotionally charged language recall failure on loaded or sarcastic tone. \\

\bottomrule
\end{tabular}
}
\caption{Representative examples of common error categories in span-level propaganda detection, including over-prediction, under-prediction, false positives, and false negatives.}
\label{Tab: error}
\end{table}

%% file: tables/humaneval.tex
\begin{table}[t]
\centering
\small
\setlength{\tabcolsep}{1pt}
\renewcommand{\arraystretch}{0.5}

\resizebox{0.5\textwidth}{!}{%

\begin{tabular}{lccc | ccc | ccc | ccc | ccc}
\toprule
\multirow{3}{*}{\textbf{Iter}} 
& \multicolumn{9}{c}{\textbf{Human Evaluation}} 
& \multicolumn{6}{c}{\textbf{Expert Evaluation}} \\

\cmidrule(lr){2-10}
\cmidrule(lr){11-16}

& \multicolumn{3}{c}{CR}
& \multicolumn{3}{c}{RT}
& \multicolumn{3}{c}{OC}
& \multicolumn{3}{c}{TP}
& \multicolumn{3}{c}{LS} \\

\cmidrule(lr){2-4}
\cmidrule(lr){5-7}
\cmidrule(lr){8-10}
\cmidrule(lr){11-13}
\cmidrule(lr){14-16}

& poor & fair & good
& poor & fair & good
& poor & fair & good
& poor & fair & good
& poor & fair & good \\
\midrule

0 & 41 & 45 & 15 & 24 & 55 & 21 & 26 & 49 & 25 & 60 & 40 & 0 & 53 & 47 & 0 \\
25      & 3  & 30 & 67 & 1  & 14 & 85 & 4  & 23 & 74 & 0  & 17 & 83 & 0  & 33 & 67 \\

\bottomrule
\end{tabular}
}
\caption{Human and expert evaluation for assessing reasoning quality(in \% values)}
\label{tab: human_expert_eval}
\end{table}

%% file: New_Tables/cross.tex
\begin{table}[h]
\centering
\small
\setlength{\tabcolsep}{3pt}

\begin{tabular}{ccccc}
\toprule
\multirow{2}{*}{\textbf{R}} &
\multicolumn{2}{c}{\textbf{PTC}} &
\multicolumn{2}{c}{\textbf{IndiProp}} \\
\cmidrule(lr){2-3}\cmidrule(lr){4-5}
& \textbf{Avg F1} & \textbf{W. Avg F1} & \textbf{Avg F1} & \textbf{W. Avg F1} \\
\midrule
1 & 0.188 & 0.196 & 0.286 & 0.270 \\
2 & 0.209 & 0.207 & 0.317 & 0.295 \\
3 & 0.185 & 0.189 & 0.327 & 0.299 \\
4 & 0.194 & 0.190 & 0.315 & 0.286 \\
5 & 0.224 & 0.205 & 0.306 & 0.279 \\
\bottomrule
\end{tabular}
\caption{Cross-dataset inference performance for the Mistral (K2) configuration. PTC was evaluated using CoT refined on IndiProp, while IndiProp was evaluated using CoT refined on PTC.}
\label{cross}
\end{table}

%% file: New_Tables/inference_time.tex
\begin{table}[t]
\centering
\scriptsize
\label{tab:timing}
\begin{tabular}{llcc}
\toprule
\textbf{Phase} & \textbf{Configuration} & \textbf{Mistral (s)} & \textbf{Qwen (s)} \\
\midrule
Inference & $R = 3$ & 6.05 & 11.70 \\
\midrule
\multirow{3}{*}{Refinement}
& $K = 1$ & 2.99 & 5.63 \\
& $K = 2$ & 4.70 & 9.01 \\
& $K = 3$ & 6.12 & 12.00 \\
\bottomrule
\end{tabular}
\caption{Inference and refinement time (in seconds) for Mistral and Qwen.}
\label{infer}
\end{table}

%% file: New_Tables/semeval_gpt.tex
\begin{table}[ht]
\centering
\scriptsize
\setlength{\tabcolsep}{5pt}
\renewcommand{\arraystretch}{0.9}
\begin{tabular}{l l cc cc}
\toprule
\multirow{2}{*}{\textbf{Config}} &
\multirow{2}{*}{\textbf{Backbone}} &
\multicolumn{2}{c}{\textbf{IndiProp}} &
\multicolumn{2}{c}{\textbf{PTC}} \\
\cmidrule(lr){3-4} \cmidrule(lr){5-6}
& & Avg & W.Avg & Avg & W.Avg \\
\midrule
0-shot & \multirow{3}{*}{GPT} & 0.231 & 0.212 & 0.114 & 0.170 \\
1-shot & & 0.176 & 0.201 & 0.170 & 0.203 \\
CoT & & 0.241 & 0.244 & 0.152 & 0.194 \\
\midrule
RePrompt & \multirow{6}{*}{GPT} & 0.106 & 0.158 & 0.140 & 0.184 \\
\model{} R1 & & 0.229 & 0.246 & 0.100 & 0.141 \\
\model{} R2 & & 0.235 & 0.251 & 0.102 & 0.141 \\
\model{} R3 & & 0.253 & 0.246 & 0.092 & 0.148 \\
\model{} R4 & & 0.264 & 0.252 & 0.130 & 0.161 \\
\model{} R5 & & 0.272 & 0.274 & 0.098 & 0.147 \\
\bottomrule
\end{tabular}
\caption{Performance comparison of the GPT backbone on the SemEval benchmark. Average F1 (Avg) and Weighted Average F1 (W.Avg) are reported for the IndiProp and PTC tasks.}
\label{tab:gpt_semeval_results}
\end{table}

%% file: New_Tables/tech_wise_indi.tex
\begin{table}[ht]
\centering
\scriptsize
\setlength{\tabcolsep}{4pt}
\renewcommand{\arraystretch}{0.9}

\label{tab:indiprop_refinement}
\begin{tabular}{lccccc}
\toprule
\textbf{Technique} & \textbf{R1} & \textbf{R2} & \textbf{R3} & \textbf{R4} & \textbf{R5} \\
\midrule
Causal Oversimplification & 0.308 & 0.291 & 0.258 & 0.208 & 0.256 \\
Black-and-white Fallacy & 0.347 & 0.384 & 0.327 & 0.371 & 0.356 \\
Straw Man & 0.280 & 0.183 & 0.269 & 0.160 & 0.291 \\
Whataboutism & 0.309 & 0.302 & 0.375 & 0.331 & 0.340 \\
Reductio ad Hitlerum & 0.299 & 0.577 & 0.625 & 0.465 & 0.625 \\
Red Herring & 0.346 & 0.483 & 0.377 & 0.523 & 0.448 \\
Loaded Language & 0.220 & 0.254 & 0.255 & 0.243 & 0.235 \\
Name Calling and Labeling & 0.378 & 0.382 & 0.368 & 0.359 & 0.381 \\
Appeal to emotion/fear/prejudice & 0.335 & 0.360 & 0.382 & 0.409 & 0.378 \\
Slogans & 0.484 & 0.486 & 0.495 & 0.412 & 0.527 \\
Flag-Waving & 0.452 & 0.471 & 0.553 & 0.511 & 0.596 \\
Exaggeration and Minimization & 0.225 & 0.245 & 0.168 & 0.223 & 0.286 \\
Thought-terminating Cliches & 0.257 & 0.319 & 0.173 & 0.216 & 0.234 \\
Bandwagon & 0.554 & 0.369 & 0.551 & 0.390 & 0.284 \\
Smears & 0.356 & 0.331 & 0.386 & 0.391 & 0.334 \\
OVC & 0.202 & 0.276 & 0.290 & 0.452 & 0.412 \\
Doubt & 0.164 & 0.240 & 0.290 & 0.301 & 0.191 \\
Appeal to Authority & 0.297 & 0.238 & 0.418 & 0.302 & 0.332 \\
Assertion & 0.381 & 0.413 & 0.399 & 0.386 & 0.421 \\
Glittering Generalities (Virtue) & 0.309 & 0.316 & 0.373 & 0.308 & 0.320 \\
\bottomrule
\end{tabular}
\caption{IndiProp per-technique F1 scores across refinement rounds (Mistral - K2(i-bias)).}
\label{indi_tech}
\end{table}

%% file: New_Tables/tab_ablation.tex
\begin{table}[t]
\centering
\resizebox{\columnwidth}{!}{
\begin{tabular}{llcccccccc}
\toprule
\multirow{2}{*}{Setup} & \multirow{2}{*}{R} 
& \multicolumn{4}{c}{Mistral} 
& \multicolumn{4}{c}{Qwen} \\
\cmidrule(lr){3-6} \cmidrule(lr){7-10}
& 
& \multicolumn{2}{c}{IndiProp} 
& \multicolumn{2}{c}{PTC}
& \multicolumn{2}{c}{IndiProp} 
& \multicolumn{2}{c}{PTC} \\
\cmidrule(lr){3-4} \cmidrule(lr){5-6} \cmidrule(lr){7-8} \cmidrule(lr){9-10}
& 
& Avg & W.Avg & Avg & W.Avg 
& Avg & W.Avg & Avg & W.Avg \\
\midrule

\multirow{5}{*}{I - BIAS}
& R=1 & 0.320 & 0.311 & 0.167 & 0.180 & 0.296 & 0.326 & 0.172 & 0.191 \\
& R=2 & 0.343 & 0.300 & 0.166 & 0.182 & 0.316 & 0.348 & 0.181 & 0.196 \\
& R=3 & 0.364 & 0.347 & 0.155 & 0.172 & 0.345 & \textbf{0.362} & 0.184 & 0.198 \\
& R=4 & 0.344 & 0.317 & 0.172 & 0.170 & 0.326 & 0.356 & \textbf{0.200} & \textbf{0.211 }\\
& R=5 & 0.361 & 0.306 & 0.172 & 0.170 & \textbf{0.359} & 0.355 & 0.170 & 0.187 \\

\midrule
\multirow{5}{*}{RANDOM}
& R=1 & 0.328 & 0.355 & 0.175 & 0.120 & 0.304 & 0.327 & 0.186 & 0.137 \\
& R=2 & 0.343 & 0.348 & 0.169 & 0.122 & 0.326 & 0.344 & 0.182 & 0.137 \\
& R=3 & 0.367 & \textbf{0.377} & 0.168 & 0.126 & 0.347 & 0.362 & 0.172 & 0.124 \\
& R=4 & 0.360 & 0.360 & 0.162 & 0.117 & 0.333 & 0.350 & 0.181 & 0.135 \\
& R=5 & \textbf{0.378} & 0.361 & 0.153 & 0.111 & 0.328 & 0.351 & 0.178 & 0.134 \\

\midrule
\multirow{5}{*}{J - BIAS}
& R=1 & 0.333 & 0.362 & 0.166 & 0.160 & 0.315 & 0.337 & 0.181 & 0.173 \\
& R=2 & 0.352 & 0.366 &  \textbf{0.188} & \textbf{0.184} & 0.306 & 0.334 & 0.184 & 0.174 \\
& R=3 & 0.343 & 0.367 & 0.171 & 0.173 & 0.319 & 0.344 & 0.170 & 0.177 \\
& R=4 & 0.331 & 0.371 & 0.174 & 0.174 & 0.326 & 0.343 & 0.177 & 0.178 \\
& R=5 & 0.360 & 0.362 & 0.161 & 0.161 & 0.328 & 0.350 & 0.176 & 0.168 \\

\bottomrule
\end{tabular}
}
\caption{\model{} performance across bias strategies with $K=2$ configuration.}
\label{tab: ablation1}

\end{table}

%% file: New_Tables/tech_wise_ptc.tex
\begin{table}[t]
\centering
\scriptsize
\setlength{\tabcolsep}{3pt}
\renewcommand{\arraystretch}{0.9}

\begin{tabular}{p{3.6cm}ccccc}
\toprule
\textbf{Technique} & \textbf{R1} & \textbf{R2} & \textbf{R3} & \textbf{R4} & \textbf{R5} \\
\midrule
Appeal to Authority & 0.128 & 0.137 & 0.147 & 0.153 & 0.128 \\
Appeal to fear/prejudice & 0.176 & 0.211 & 0.173 & 0.195 & 0.202 \\
Bandwagon, Reductio ad Hitlerum & 0.253 & 0.167 & 0.167 & 0.167 & 0.167 \\
Black-and-white Fallacy & 0.047 & 0.106 & 0.089 & 0.059 & 0.084 \\
Causal Oversimplification & 0.182 & 0.157 & 0.111 & 0.224 & 0.197 \\
Doubt & 0.142 & 0.194 & 0.189 & 0.262 & 0.218 \\
Exaggeration and Minimization & 0.133 & 0.086 & 0.087 & 0.097 & 0.097 \\
Flag Waving & 0.173 & 0.256 & 0.223 & 0.225 & 0.281 \\
Loaded Language & 0.189 & 0.194 & 0.180 & 0.174 & 0.157 \\
Name Calling and Labeling & 0.281 & 0.266 & 0.280 & 0.274 & 0.276 \\
Repetition & 0.172 & 0.116 & 0.108 & 0.105 & 0.110 \\
Slogans & 0.133 & 0.147 & 0.125 & 0.130 & 0.133 \\
Thought-terminating Cliches & 0.136 & 0.074 & 0.075 & 0.053 & 0.126 \\
Whataboutism, Strawmen and Red Herring & 0.197 & 0.209 & 0.219 & 0.221 & 0.174 \\
\bottomrule
\end{tabular}

\caption{PTC per-technique F1 scores across refinement rounds (Mistral - k2(i-bias)).}
\label{ptc_tech}

\end{table}

%% file: tables/k1_ablation.tex
\begin{table*}[t]
\centering
\scriptsize
\setlength{\tabcolsep}{3pt} 
\begin{tabular}{lllcccccc|cccccc}
\toprule
\multirow{3}{*}{Setup} & \multirow{3}{*}{Backbone} & \multirow{3}{*}{Config}
& \multicolumn{6}{c}{IndiProp}
& \multicolumn{6}{c}{PTC} \\

\cmidrule(lr){4-9} \cmidrule(lr){10-15}

& & 
& \multicolumn{2}{c}{$R_{\text{LCS}}$}
& \multicolumn{2}{c}{$P_{\text{LCS}}$}
& \multicolumn{2}{c}{$F1_{\text{LCS}}$}
& \multicolumn{2}{c}{$R_{\text{LCS}}$}
& \multicolumn{2}{c}{$P_{\text{LCS}}$}
& \multicolumn{2}{c}{$F1_{\text{LCS}}$} \\

\cmidrule(lr){4-5} \cmidrule(lr){6-7} \cmidrule(lr){8-9}
\cmidrule(lr){10-11} \cmidrule(lr){12-13} \cmidrule(lr){14-15}

& & 
& Avg & W.Avg & Avg & W.Avg & Avg & W.Avg
& Avg & W.Avg & Avg & W.Avg & Avg & W.Avg \\
\midrule

\multirow{2}{*}{SFT}
& Hitachi & -- 
& 0.381 & 0.438 & 0.150 & 0.201 & 0.180 & 0.232
& 0.179 & 0.217 & 0.084 & 0.108 & 0.093 & 0.118 \\
& Aschern & -- 
& 0.311 & 0.345 & 0.203 & 0.236 & 0.205 & 0.231
& 0.248 & 0.296 & 0.206 & 0.219 & 0.174 & 0.203 \\

\midrule
\multirow{4}{*}{Prompting}
& \multirow{4}{*}{Mistral} & 0-shot 
& 0.270 & 0.276 & 0.319 & 0.345 & 0.263 & 0.273
& 0.259 & 0.243 & 0.199 & 0.211 & 0.187 & 0.184 \\
& & 1-shot 
& 0.365 & 0.346 & 0.340 & 0.365 & 0.318 & 0.317
& 0.244 & 0.215 & 0.214 & 0.225 & 0.200 & 0.192 \\
& & 2-shot 
& 0.154 & 0.157 & 0.154 & 0.157 & 0.154 & 0.157
& 0.250 & 0.225 & 0.218 & 0.237 & 0.195 & 0.194 \\
& & CoT 
& 0.343 & 0.352 & 0.294 & 0.320 & 0.285 & 0.300
& 0.244 & 0.253 & 0.187 & 0.210 & 0.186 & 0.202 \\

\midrule
\multirow{2}{*}{Reasoning}
& \multirow{2}{*}{Mistral} & RePrompting 
& 0.382 & 0.369 & 0.352 & 0.349 & 0.326 & 0.314
& 0.258 & 0.258 & 0.158 & 0.197 & 0.160 & 0.186 \\
& & EvoPrompt 
& 0.256 & 0.306 & 0.217 & 0.227 & 0.213 & 0.227
& 0.202 & 0.242 & 0.182 & 0.204 & 0.171 & 0.195 \\

\midrule
\multirow{4}{*}{Prompting}
& \multirow{4}{*}{GPT} & 0-shot 
& 0.314 & 0.285 & 0.345 & 0.355 & 0.294 & 0.279
& 0.194 & 0.173 & 0.197 & 0.204 & 0.169 & 0.163 \\
& & 1-shot 
& 0.249 & 0.253 & 0.320 & 0.356 & 0.254 & 0.263
& 0.259 & 0.258 & 0.237 & 0.239 & 0.218 & 0.215 \\
& & 2-shot 
& 0.269 & 0.267 & 0.341 & 0.371 & 0.275 & 0.279
& 0.278 & 0.261 & 0.226 & 0.223 & 0.212 & 0.203 \\
& & CoT 
& 0.313 & 0.297 & 0.361 & 0.364 & 0.302 & 0.289
& 0.265 & 0.257 & 0.234 & 0.231 & 0.210 & 0.208 \\

\midrule
\multirow{2}{*}{Reasoning}
& \multirow{2}{*}{GPT} & RePrompting 
& 0.306 & 0.312 & 0.287 & 0.291 & 0.297 & 0.308
& 0.228 & 0.234 & 0.216 & 0.246 & 0.193 & 0.209 \\
& & EvoPrompt 
& 0.484 & 0.515 & 0.237 & 0.262 & 0.272 & 0.300
& 0.419 & 0.366 & 0.200 & 0.178 & 0.232 & 0.208 \\
\midrule
\midrule
\multirow{15}{*}{$K=1$}
& \multirow{5}{*}{Mistral}
& ReCoT R1 & 0.381 & 0.377 & 0.330 & 0.367 & 0.323 & 0.336 & 0.227 & 0.211 & 0.185 & 0.220 & 0.171 & 0.185 \\
& & ReCoT R2 & 0.400 & 0.390 & 0.348 & 0.360 & 0.340 & 0.337 & 0.232 & 0.225 & 0.184 & 0.218 & 0.171 & 0.185 \\
& & ReCoT R3 & 0.447 & 0.447 & 0.349 & 0.370 & 0.354 & 0.360 & 0.234 & 0.229 & 0.200 & 0.236 & 0.182 & 0.196 \\
& & ReCoT R4 & 0.402 & 0.413 & 0.337 & 0.362 & 0.334 & 0.345 & 0.242 & 0.236 & 0.211 & 0.245 & 0.193 & 0.205 \\
& & ReCoT R5 & 0.424 & 0.427 & 0.364 & 0.379 & 0.357 & 0.360 & 0.236 & 0.229 & 0.200 & 0.230 & 0.186 & 0.196 \\

\cmidrule(lr){2-15}
& \multirow{5}{*}{Qwen}
& ReCoT R1 & 0.356 & 0.366 & 0.340 & 0.371 & 0.323 & 0.339 & 0.233 & 0.230 & 0.194 & 0.228 & 0.184 & 0.201 \\
& & ReCoT R2 & 0.338 & 0.372 & 0.335 & 0.376 & 0.313 & 0.344 & 0.230 & 0.235 & 0.201 & 0.230 & 0.190 & 0.205 \\
& & ReCoT R3 & 0.368 & 0.398 & 0.331 & 0.380 & 0.321 & 0.358 & 0.230 & 0.234 & 0.199 & 0.222 & 0.187 & 0.199 \\
& & ReCoT R4 & 0.385 & 0.390 & 0.351 & 0.375 & 0.341 & 0.352 & 0.217 & 0.222 & 0.192 & 0.222 & 0.177 & 0.192 \\
& & ReCoT R5 & 0.368 & 0.377 & 0.340 & 0.369 & 0.323 & 0.342 & 0.211 & 0.220 & 0.182 & 0.207 & 0.170 & 0.185 \\

\cmidrule(lr){2-15}
& \multirow{5}{*}{GPT}
& ReCoT R1 & 0.313 & 0.299 & 0.333 & 0.345 & 0.296 & 0.289 & 0.201 & 0.207 & 0.209 & 0.238 & 0.178 & 0.194 \\
& & ReCoT R2 & 0.315 & 0.326 & 0.323 & 0.351 & 0.290 & 0.304 & 0.204 & 0.218 & 0.185 & 0.215 & 0.169 & 0.190 \\
& & ReCoT R3 & 0.312 & 0.322 & 0.320 & 0.333 & 0.287 & 0.293 & 0.202 & 0.217 & 0.172 & 0.199 & 0.166 & 0.185 \\
& & ReCoT R4 & 0.330 & 0.333 & 0.332 & 0.345 & 0.307 & 0.307 & 0.210 & 0.220 & 0.185 & 0.215 & 0.172 & 0.191 \\
& & ReCoT R5 & 0.328 & 0.324 & 0.326 & 0.337 & 0.301 & 0.299 & 0.207 & 0.218 & 0.184 & 0.217 & 0.171 & 0.192 \\
\midrule
\midrule

\multirow{15}{*}{$K=2$}
& \multirow{5}{*}{Mistral}
& ReCoT R1 & 0.380 & 0.351 & 0.320 & 0.321 & 0.320 & 0.311 & 0.241 & 0.223 & 0.170 & 0.202 & 0.167 & 0.180 \\
& & ReCoT R2 & 0.404 & 0.343 & 0.349 & 0.311 & 0.343 & 0.300 & 0.239 & 0.227 & 0.171 & 0.206 & 0.166 & 0.182 \\
& & ReCoT R3 & 0.457 & 0.446 & 0.354 & 0.330 & 0.364 & 0.347 & 0.224 & 0.216 & 0.161 & 0.194 & 0.155 & 0.172 \\
& & ReCoT R4 & 0.420 & 0.375 & 0.334 & 0.318 & 0.344 & 0.317 & 0.239 & 0.216 & 0.170 & 0.183 & 0.172 & 0.170 \\
& & ReCoT R5 & 0.429 & 0.364 & 0.357 & 0.303 & 0.361 & 0.306 & 0.243 & 0.215 & 0.172 & 0.183 & 0.172 & 0.170 \\

\cmidrule(lr){2-15}
& \multirow{5}{*}{Qwen}
& ReCoT R1 & 0.338 & 0.357 & 0.311 & 0.353 & 0.296 & 0.326 & 0.210 & 0.216 & 0.180 & 0.217 & 0.172 & 0.191 \\
& & ReCoT R2 & 0.345 & 0.375 & 0.334 & 0.380 & 0.316 & 0.348 & 0.222 & 0.225 & 0.191 & 0.221 & 0.181 & 0.196 \\
& & ReCoT R3 & 0.388 & 0.401 & 0.352 & 0.387 & 0.345 & 0.362 & 0.233 & 0.233 & 0.194 & 0.222 & 0.184 & 0.198 \\
& & ReCoT R4 & 0.363 & 0.393 & 0.343 & 0.379 & 0.326 & 0.356 & 0.256 & 0.250 & 0.209 & 0.233 & 0.200 & 0.211 \\
& & ReCoT R5 & 0.406 & 0.396 & 0.368 & 0.377 & 0.359 & 0.355 & 0.226 & 0.225 & 0.175 & 0.204 & 0.170 & 0.187 \\

\cmidrule(lr){2-15}
& \multirow{5}{*}{GPT}
& ReCoT R1 & 0.342 & 0.225 & 0.347 & 0.243 & 0.322 & 0.215 & 0.218 & 0.151 & 0.184 & 0.144 & 0.177 & 0.128 \\
& & ReCoT R2 & 0.325 & 0.231 & 0.321 & 0.236 & 0.297 & 0.214 & 0.196 & 0.138 & 0.187 & 0.149 & 0.168 & 0.124 \\
& & ReCoT R3 & 0.358 & 0.248 & 0.336 & 0.229 & 0.320 & 0.216 & 0.212 & 0.154 & 0.175 & 0.147 & 0.165 & 0.130 \\
& & ReCoT R4 & 0.362 & 0.248 & 0.318 & 0.224 & 0.313 & 0.215 & 0.202 & 0.145 & 0.181 & 0.152 & 0.164 & 0.128 \\
& & ReCoT R5 & 0.341 & 0.245 & 0.314 & 0.228 & 0.301 & 0.215 & 0.183 & 0.138 & 0.175 & 0.148 & 0.160 & 0.127 \\

\midrule
\midrule

\multirow{15}{*}{$K=3$}
& \multirow{5}{*}{Mistral}
& ReCoT R1 & 0.436 & 0.410 & 0.341 & 0.340 & 0.347 & 0.333 & 0.231 & 0.193 & 0.155 & 0.175 & 0.156 & 0.153 \\
& & ReCoT R2 & 0.394 & 0.412 & 0.316 & 0.346 & 0.319 & 0.340 & 0.218 & 0.195 & 0.162 & 0.177 & 0.157 & 0.157 \\
& & ReCoT R3 & 0.438 & 0.478 & 0.325 & 0.323 & 0.334 & 0.338 & 0.237 & 0.210 & 0.166 & 0.176 & 0.168 & 0.165 \\
& & ReCoT R4 & 0.432 & 0.479 & 0.307 & 0.309 & 0.324 & 0.331 & 0.230 & 0.199 & 0.152 & 0.167 & 0.153 & 0.152 \\
& & ReCoT R5 & 0.437 & 0.478 & 0.308 & 0.331 & 0.328 & 0.347 & 0.226 & 0.201 & 0.152 & 0.167 & 0.156 & 0.156 \\

\cmidrule(lr){2-15}
& \multirow{5}{*}{Qwen}
& ReCoT R1 & 0.371 & 0.382 & 0.345 & 0.367 & 0.330 & 0.343 & 0.232 & 0.229 & 0.208 & 0.233 & 0.194 & 0.204 \\
& & ReCoT R2 & 0.345 & 0.375 & 0.332 & 0.373 & 0.310 & 0.343 & 0.224 & 0.229 & 0.197 & 0.229 & 0.183 & 0.199 \\
& & ReCoT R3 & 0.396 & 0.419 & 0.345 & 0.364 & 0.340 & 0.355 & 0.218 & 0.219 & 0.189 & 0.215 & 0.178 & 0.192 \\
& & ReCoT R4 & 0.387 & 0.418 & 0.331 & 0.347 & 0.326 & 0.343 & 0.209 & 0.213 & 0.174 & 0.203 & 0.166 & 0.182 \\
& & ReCoT R5 & 0.410 & 0.425 & 0.353 & 0.365 & 0.343 & 0.353 & 0.227 & 0.217 & 0.184 & 0.198 & 0.180 & 0.183 \\

\cmidrule(lr){2-15}
& \multirow{5}{*}{GPT}
& ReCoT R1 & 0.347 & 0.186 & 0.312 & 0.180 & 0.305 & 0.168 & 0.204 & 0.179 & 0.193 & 0.200 & 0.175 & 0.166 \\
& & ReCoT R2 & 0.328 & 0.187 & 0.297 & 0.168 & 0.285 & 0.162 & 0.188 & 0.177 & 0.171 & 0.186 & 0.155 & 0.158 \\
& & ReCoT R3 & 0.341 & 0.194 & 0.290 & 0.168 & 0.286 & 0.162 & 0.188 & 0.179 & 0.159 & 0.174 & 0.149 & 0.153 \\
& & ReCoT R4 & 0.334 & 0.191 & 0.295 & 0.170 & 0.287 & 0.163 & 0.195 & 0.183 & 0.169 & 0.183 & 0.157 & 0.159 \\
& & ReCoT R5 & 0.348 & 0.194 & 0.303 & 0.174 & 0.298 & 0.168 & 0.196 & 0.187 & 0.161 & 0.177 & 0.154 & 0.159 \\

\bottomrule
\end{tabular}

\caption{$F1_{LCS}$, $R_{LCS}$, and $P_{LCS}$ for \model{} with $I_{bias}$ at $K \in \{1, 2, 3\}$ across Mistral, Qwen, and GPT on \dataset{} and \PTC{}}
\label{tab:k1_k3_ablation}

\end{table*}

%% file: tables/cot_compare.tex
\begin{table*}[t]
\centering
\scriptsize %
\renewcommand{\arraystretch}{1.4} 

\resizebox{\textwidth}{!}{
\begin{tabular}{lp{0.95\textwidth}} %
\toprule
\textbf{Variant} & \textbf{Evaluation Details} \\
\midrule

\textbf{Initialization ($n=0$)} &  
<CoT> \newline  
Step 1: Identify the main arguments presented in the article. The article discusses the contrasting actions of citizens during a public appreciation event for frontline workers and the protests against CAA, NRC, and NPR led by Suranya Aiyar. \newline 
Step 2: Look for instances where an opposing viewpoint is misrepresented. The article portrays Suranya Aiyar's protests as being out of touch or inappropriate given the context of a pandemic and the collective appreciation of healthcare workers, suggesting that those who protest are disregarding the situation at hand. \newline  
Step 3: Analyze specific statements that may reduce the opposing viewpoint to an unfairly simple or easily attackable position. The article claims "partisan political interests even in the face of an apocalypse." This implies that any dissent against the government's practices during a crisis is wrong or irrelevant, which misrepresents the nuanced argument that some may hold against these policies, portraying them as entirely neglectful of the situation. \newline 
Step 4: The statement about residents of Jamia Nagar raising slogans while others rallied for healthcare workers further simplifies their stance to mere opposition, not recognizing the validity of their protest or their right to express it.  \newline  
</CoT> \newline\newline
\textbf{Extracted Span(s):} \newline
1. ``some people would not abandon their partisan political interests even in the face of an apocalypse.'' \newline
2. ``On a similar note, when the nation demonstrated their appreciation towards those providing people with healthcare in times of this major crisis, residents of Jamia Nagar's Abul Fazal colony raised `Azaadi' slogans instead.'' \newline\newline
\textbf{F1 Score:} 0.0841 \\

\midrule

\textbf{I-Bias} ($n=25$) &  
<CoT> \newline  
Step 1: The article describes the actions of Suranya Aiyar and others during a nationwide ``Janta Curfew'' to show appreciation for frontline workers battling the pandemic. \newline  
Step 2: Suranya Aiyar is portrayed as screaming slogans demanding the rollback of CAA, NRC, and NPR, which are contentious issues in India. \newline  
Step 3: The article criticizes her actions by stating that she ``riled against the people as the nation came together as one,'' implying that her actions were disruptive and inappropriate during a time of national unity. \newline  
Step 4: To identify Straw Man propaganda, we need to check if the article misrepresents or exaggerates Suranya Aiyar's actions or intentions. The statement ``riled against the people as the nation came together as one'' could be interpreted as a Straw Man if it attributes an extreme or simplistic viewpoint to her actions that may not be accurate. \newline  
Conclusion: Propaganda spans exist \newline  
</CoT> \newline  
<span>riled against the people as the nation came together as one</span> \newline\newline
\textbf{Extracted Span(s):} ``riled against the people as the nation came together as one'' \newline\newline
\textbf{F1 Score:} 0.7060 \\

\midrule

\textbf{J-Bias ($n=25$)} &  
<CoT> \newline  
Step 1: To determine if there are any ``Straw Man'' arguments in the article, I need to identify whether the article misrepresents or oversimplifies the position of any opposing side or group. A Straw Man argument often involves refuting a weaker version of the actual argument presented by the opposing side. \newline  
Step 2: Analyzing the article, it describes the actions of Suranya Aiyar and other individuals who raised slogans during the `Janta Curfew'. The article portrays these actions as disruptive and partisan, contrasting them with the nationwide effort to show appreciation for frontline workers battling the pandemic. \newline  
Step 3: I look for any indication that the article misrepresents the motivations or arguments of Suranya Aiyar or the other individuals mentioned. The article states that Suranya Aiyar and others ``riled against the people as the nation came together as one,'' and that their actions demonstrate a refusal to abandon ``partisan political interests even in the face of an apocalypse.'' This could be seen as a misrepresentation or oversimplification of their actions, as it frames their political statements as inherently disruptive and selfish, without exploring their underlying motivations or the context of their slogans. \newline  
Conclusion: Propaganda spans exist. \newline  
</CoT> \newline  
<span>riled against the people as the nation came together as one</span> \newline  
<span>some people would not abandon their partisan political interests even in the face of an apocalypse</span> \newline\newline
\textbf{Extracted Span(s):} \newline
1. ``riled against the people as the nation came together as one'' \newline
2. ``some people would not abandon their partisan political interests even in the face of an apocalypse'' \newline\newline
\textbf{F1 Score:} 0.5500 \\
\midrule

\textbf{Random ($n=25$)} &  
<CoT> \newline  
Step 1: To identify if there are any statements that qualify as straw man propaganda, I need to check if the article misrepresents an opponent's position in order to make it easier to attack or refute. \newline  
Step 2: The article discusses Suranya Aiyar's actions and slogans during the `Janta Curfew'. It mentions her demands for the rollback of CAA, NRC, and NPR, and her slogans in support of various causes. \newline  
Step 3: The article states, ``Suranya Aiyar was walking around her lawn presumably while she raised these slogans at 5 pm on Sunday.'' This could be interpreted as a misrepresentation if the context or intent behind her actions is not accurately portrayed. Additionally, the phrase ``some people would not abandon their partisan political interests even in the face of an apocalypse'' could be seen as a straw man argument if it oversimplifies or misrepresents the motivations of those opposing CAA, NRC, and NPR. \newline  
Conclusion: Propaganda spans exist. \newline  
</CoT> \newline  
<span>some people would not abandon their partisan political interests even in the face of an apocalypse.</span> \newline\newline
\textbf{Extracted Span(s):} ``some people would not abandon their partisan political interests even in the face of an apocalypse.'' \newline\newline
\textbf{F1 Score:} 0.1000 \\
\bottomrule
\end{tabular}}
\caption{Propaganda Span and Reasoning (CoT) Results}
\label{tab: propaganda_spans}
\end{table*}